\documentclass[10pt,twocolumn,letterpaper]{article}

\usepackage[pagenumbers]{cvpr} 

\definecolor{cvprblue}{rgb}{0.21,0.49,0.74}
\usepackage[pagebackref,breaklinks,colorlinks,allcolors=cvprblue]{hyperref}
\usepackage{bm}
\usepackage{mathrsfs}
\usepackage{multirow, multicol}
\usepackage{graphicx}
\usepackage{colortbl}
\usepackage{adjustbox}

\newcommand{\methodname}[1]{BlobGEN-Vid}

\def\paperID{10628} 
\def\confName{CVPR}
\def\confYear{2025}

\title{BlobGEN-Vid: Compositional Text-to-Video Generation with Blob Video Representations}

\author{{Weixi Feng\textsuperscript{1}\quad\quad Chao Liu\textsuperscript{2}\quad\quad Sifei Liu$\textsuperscript{2}$\quad\quad William Yang Wang\textsuperscript{1}} \\[0.1cm]
{Arash Vahdat\textsuperscript{2}\quad\quad\quad Weili Nie\textsuperscript{2}}\\[0.3cm]
{\textsuperscript{1}UC Santa Barbara} \quad {\textsuperscript{2}NVIDIA}\\[0.3cm]
\href{https://blobgen-vid2.github.io/}{blobgen-vid2.github.io}
}

\begin{document}

\twocolumn[{
\renewcommand\twocolumn[1][]{#1}%
\maketitle
\vspace{-0.5cm}
\includegraphics[width=0.95\textwidth]{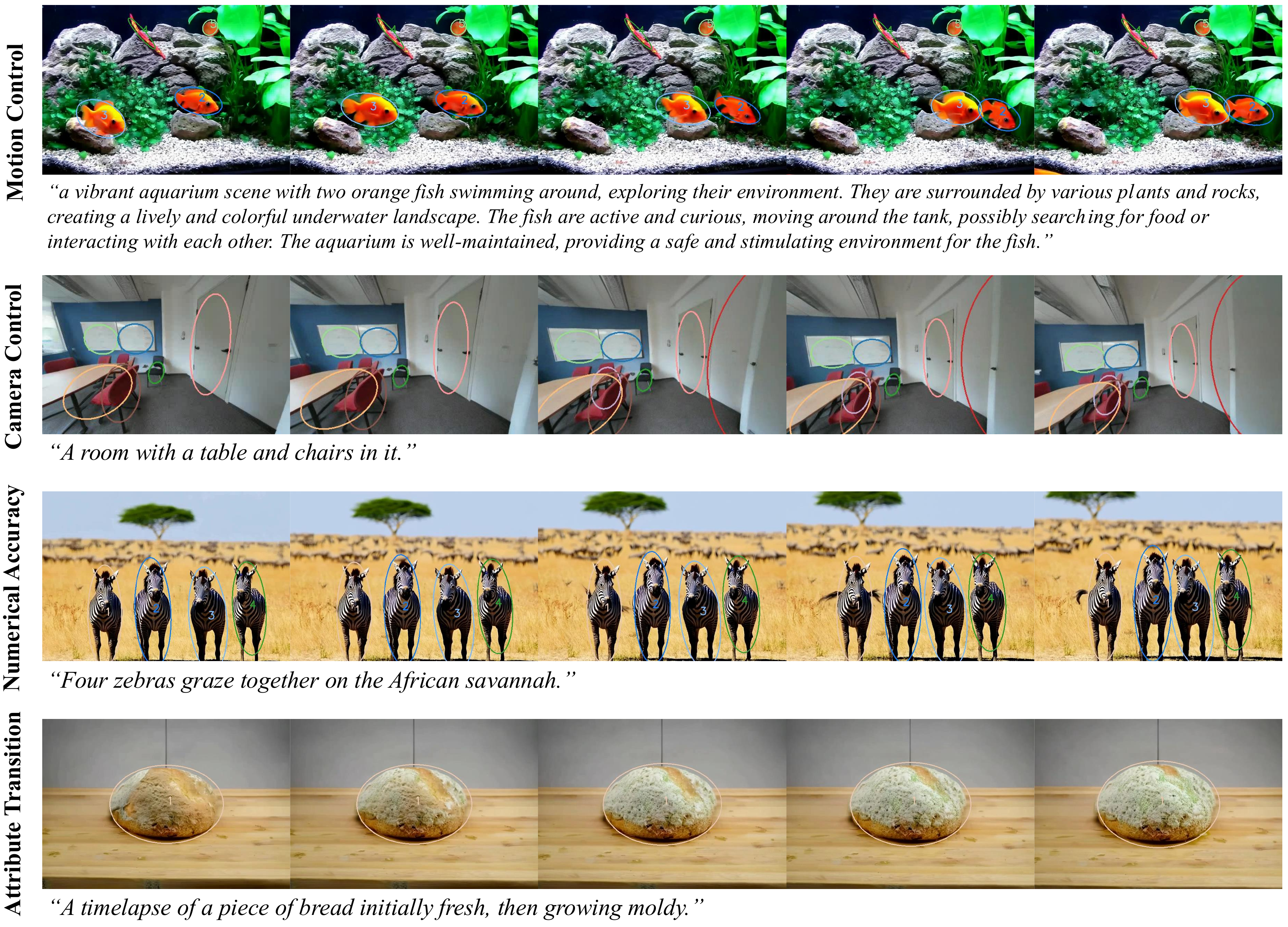}
\captionof{figure}{With blob video representations, \methodname{} can support fine-grained controllability in text-to-video generation in terms of motion control, camera control, numerical accuracy and attribute transition. Blobs in top two rows are extracted from a video and a 3D scene, respectively, using the pre-trained segmentation model and image captioning model, while blobs in bottom two rows are generated by GPT-4o with the given global prompt as input.} 
\vspace{2em}
\label{fig:teaser}
}]

\begin{abstract}
Existing video generation models struggle to follow complex text prompts and synthesize multiple objects, raising the need for additional grounding input for improved controllability. In this work, we propose to decompose videos into visual primitives -- blob video representation, a general representation for controllable video generation. Based on blob conditions, we develop a blob-grounded video diffusion model named \methodname{} that allows users to control object motions and fine-grained object appearance. In particular, we introduce a masked 3D attention module that effectively improves regional consistency across frames. In addition, we introduce a learnable module to interpolate text embeddings so that users can control semantics in specific frames and obtain smooth object transitions. We show that our framework is model-agnostic and build \methodname{} based on both U-Net and DiT-based video diffusion models. Extensive experimental results show that \methodname{} achieves superior zero-shot video generation ability and state-of-the-art layout controllability on multiple benchmarks. When combined with an LLM for layout planning, our framework even outperforms proprietary text-to-video generators in terms of compositional accuracy. 
\end{abstract} 
\section{Introduction}
\label{sec:intro}

Recent advancements in text-to-video generation have enabled us to generate more realistic videos with high visual quality and intricate motions. These advancements are driven by new model architectures \cite{blattmann2023align, hongcogvideo, yang2024cogvideox}, improved training techniques \cite{blattmann2023stable, chen2024videocrafter2, li2024t2v} and large-scale video datasets \cite{xue2022advancing, chen2024panda}. Despite the progress, existing text-to-video models still struggle to follow complex prompts, where they often neglect key objects or confuse multiple objects as one concept. In addition, users cannot control semantic transitions or camera motion with merely text descriptions with these models. Therefore, it remains an open challenge to enhance the compositionality and controllability of video generators with layout guidance in the diffusion process. 

To resolve these challenges, recent studies propose to condition video diffusion models on visual layouts. Since a text prompt can be ambiguous in object locations and visual appearances, video generators often fail to generate scenes with large motion or complex compositions. Additional grounding inputs can guide the generation process for enhanced controllability. These layouts are usually represented by bounding boxes moving across frames \cite{lianllm, lin2023videodirectorgpt, li2023trackdiffusion, wangboximator}. Compared to other modalities such as depth \cite{rombach2022high} or semantic maps \cite{zhang2023adding}, bounding boxes are easier to create and manipulate by users while providing coarse-grained information of local objects. However, 2D bounding boxes lack perspective invariance: the 3D counterpart of a 2D bounding box on an image is not a 3D bounding box and vice versa. This makes it difficult to synthesize 3D scenes using models grounded by bounding-boxes.

In this work, we introduce a new type of visual layouts for video generation, named \textit{blob video representations}, to serve as grounding conditions. Each blob sequence corresponds to an object instance and can be automatically extracted from videos (or 3D scenes), making it a more general and robust representation for different visual domains. Specifically, a blob video representation has two components: 1) the blob parameters, which formulate a tilted ellipse to specify the object's location, size, and orientation; and 2) the blob description, which is a free-form language description of the object's visual attributes. With this definition, our blob representation enables both motion and semantic control of visual compositions. It is also convenient for users to create and manipulate such representations as the blob parameters can be represented as structured text. 

While layout conditions have been widely studied in image generation \cite{li2023gligen, zhang2023adding, chen2024training, niecompositional}, directly applying these methods in video can lead to temporal inconsistency or compromised layout control~\cite{li2023trackdiffusion}. Some recent studies have adapted these conditions for video generation with new techniques \cite{li2023trackdiffusion, wangboximator}. However, they still suffer from the above issues and are limited to class conditions for each object box. To this end, we develop a blob-grounded text-to-video diffusion framework, termed \methodname{}, that is built upon existing video diffusion models using blob representations as grounding input. In our framework, we introduce a masked 3D attention module that facilitates object-centric spatial-temporal attention. We also utilize masked cross-attentions \cite{niecompositional} to fuse free-form object descriptions into the blob regions. As some frames do not have blob captions, we integrate a context interpolation module to enhance semantic transition throughout time.

\methodname{} is a model-agnostic framework that can be applied to both UNet~\citep{ho2020denoising} and DiT~\citep{peebles2023scalable} based diffusion models. Our experiments in open-domain video generation indicate that \methodname{} outperforms existing layout-guided video generators by a large margin in multiple dimensions. We evaluate \methodname{} on a wide range of benchmarks \cite{xu2016msr} and show that it improves the layout controllability by at least 20\% in mIOU and prompt alignment by 5\% in CLIP similarity. When combined with a large language model (LLMs) for blob planning, our pipeline outperforms proprietary video generators in mutiple aspects. Last but not least, we demonstrate that \methodname{} also achieves improved consistency and camera control in multi-view image generation in indoor scenes. 

\textbf{Our contributions}: \emph{(i)} We propose a new blob representation for text-to-video generation that enables fine-grained control of each object such as its motion and appearance. \emph{(ii)} We propose \methodname{}, a blob-grounded framework that incorporates two types of masked attention modules and a context interpolation module to pre-trained video diffusion models for regional control and temporal consistency. \methodname{} can be applied to both UNet and DiT based diffusion backbones. \emph{(iii)} We conduct extensive experiments in open-domain video generation and multi-view indoor scene generation, demonstrating \methodname{}'s superior object-level controllability and temporal consistency in generating high-quality videos.

\section{Related work}\label{sec:related}

\paragraph{Text-to-video generation.} The field of text-to-video generation (T2V) has gained much attention thanks to the advancement in new model architecture \cite{blattmann2023align, peebles2023scalable, opensora, opensora-plan}, large-scale video datasets \cite{xue2022advancing, chen2024panda}, and improved training techniques \cite{chen2024videocrafter2, esser2024scaling, li2024t2v}. There are mainly two streams of video diffusion models in terms of model architecture. Primitive video diffusion models such as Video LDM \cite{blattmann2023align}, VideoCrafter \cite{chen2023videocrafter1, chen2024videocrafter2} and some others \cite{wang2023modelscope, wang2023lavie, zhang2024show} are achieved by adding temporal self-attention modules into a U-Net diffusion backbone \cite{ho2020denoising}. The U-Net usually operates in the latent space \cite{vahdat2021score} and can be inherited from a pre-trained text-to-image model such as Stable Diffusion \cite{blattmann2023align}. Recently, as the scale of model and dataset increases, a few models based on Diffusion Transformers (DiT) \cite{peebles2023scalable} have been proposed. For example, CogVideoX \cite{yang2024cogvideox} proposes an expert DiT with stacked 3D attention blocks working on the concatenation of context embeddings and visual tokens.  As synthetic videos are becoming more realistic, it is essential to endow controllability to the generation process. 

\paragraph{Compositional video generation.} While compositional generation have been extensively studied in the image domain~\citep{li2023gligen,niecompositional,Liu2024BlobGen3D}, the compositional tasks in video generation still demand more focus~\cite{yang2024compositional, tian2024videotetris}. 
Recently, a few works start to tackle compositional video generation. Vico \cite{niecompositional} regularizes text tokens' attention maps to improve scene correctness with multiple objects. VideoTetris \cite{tian2024videotetris} proposes a spatial-temporal composing mechanism to deal with compositional change in long video generation. 
Several benchmarks are proposed to characterize compositionality \cite{liu2024fetv, liu2024evalcrafter}. Beyond issues carried from image generation, the temporal dimension in videos introduces new challenging problems. For example, T2V-CompBench \cite{sun2024t2v} and TC-Bench \cite{feng2024tc} features dynamic binding relations or object status change. These benchmarks show that existing T2V models lack of robust compositionality in generating videos with complex scenes and motions. 

\paragraph{Layout-guided video generation.} There are several studies that attempt to add layout condition on top of a pre-trained T2V model. TrackDiffusion \cite{li2023trackdiffusion} inserts trainable gated cross attentions with an instance enhancer to improve object consistency across frames. Boximator \cite{wangboximator} applies a self-attention to fuse object category and box coordinates into visual tokens. It also proposes self-tracking technique that fine-tunes the model to generate visible bounding boxes around objects first and then forget the behavior. LVD \cite{lianllm} and VideoDirectorGPT \cite{lin2023videodirectorgpt} adopt LLMs to plan bounding boxes for several keyframes, which are then passed to a video generator. As shown later, these methods may suffer from inconsistency issue and lack of controllabbility with complex layouts. 
\section{Preliminary: BlobGEN}
\label{sec:prelim_blobgen}

BlobGEN~\citep{niecompositional} first introduced blob representations to guide the open-domain image generation. It has shown that blobs can provide more fine-grained controllability than other visual layouts (such as bounding boxes) in previous layout-conditioned approaches~\citep{li2023gligen,feng2024layoutgpt}, which motivates us to use blob representations for video generation. We will next introduce the blob representations and key method design in BlobGEN, which our method is built upon.

\paragraph{Blob representations.} 
The blob representations denote the object-level visual primitives in a scene, each of which consists of two components: blob parameters and blob description. The blob parameters depict an object's shape, size and location with a vector of five variables $\tau = [c_x, c_y, a,b ,\theta]$ that defines a tilted ellipse, where $(c_x, c_y)$ is the center point of the ellipse, $a, b$ are the radii of its semi-major and semi-minor axes, and $\theta\in (-\pi, \pi]$ is the orientation angle of the ellipse. The blob description denoted by $s$ captures the visual appearance of an object using a region-level synthetic caption extracted by an image captioning model. Compared with other visual layouts (such as boxes and semantic maps), blob representations have both two advantages: 1) they retain the fine-grained spatial and appearance information about the objects in a complex scene; and 2) they can be easily constructed and manipulated by either human users or LLMs with in-context learning, since they are essentially in the form of text sentences~\citep{niecompositional}. 

To encode blob representations into blob embeddings in BlobGEN, we first obtain the blob parameter embedding $\pmb{e}_\tau \in \mathbb{R}^{\frac{d}{2}}$ with Fourier feature encoding~\cite{tancik2020fourier} and the blob description embedding $\pmb{e}_s := [\pmb{e}_{s_1}, \cdots, \pmb{e}_{s_L}] \in \mathbb{R}^{L \times \frac{d}{2}}$ with CLIP text encoder, separately, where $\frac{d}{2}$ denotes the embedding feature size and $L$ denotes the sentence length of blob description. 
We then concatenate $\pmb{e}_{\tau}$ with each $\pmb{e}_{s_l}$ along the embedding feature dimension to get $\tilde{\pmb{e}}_l:=[\pmb{e}_{\tau};\pmb{e}_{s_l}]\in \mathbb{R}^{d}$ and feed it into an MLP network to get the final blob embeddings $\pmb{e}_{\text{blob}}=\text{MLP}([\tilde{\pmb{e}}_1,...,\tilde{\pmb{e}}_L]) \in \mathbb{R}^{L\times d}$.

\paragraph{Masked cross-attention.} To incorporate the blob representations into the existing text-to-image models, BlobGEN adopts the similar network design of GLIGEN~\citep{li2023gligen} and introduces new masked cross-attention layers in a
gated way. 
Specifically, in the masked cross-attention layer, each blob embedding only attends to visual features in its local region as the visual feature maps are
masked by the (rescaled) blob ellipses.
Assume there are $N$ blobs in an image, and the visual feature map is denoted by $\pmb{g} \in \mathbb{R}^{hw \times d_g}$, where $h$ and $w$ represent the spatial size of the visual features map. We use $n$ as the index for the $n^{th}$ blob, and define the query, key and value (with different linear projections) for cross-attention as $\pmb{q} := \pmb{g}\pmb{W}_q \in \mathbb{R}^{hw \times d_g}$, $\pmb{k}^{(n)} := \pmb{e}_{\text{blob}}^{(n)}\pmb{W}_k^{(n)} \in \mathbb{R}^{L \times d_g}$, and $\pmb{v}^{(n)} := \pmb{e}_{\text{blob}}^{(n)}\pmb{W}_v^{(n)} \in \mathbb{R}^{L \times d_g}$, respectively.
The masked cross-attention is defined as
\begin{align}\label{maskca}
\small
    \text{MaskCA} := \text{Softmax} \left(\frac{[\pmb{a}^{(1)}; \cdots; \pmb{a}^{(N)}]}{\sqrt{d_g}} \right) [\pmb{v}^{(1)}; \cdots; \pmb{v}^{(N)}]
\end{align}
where the $n^{th}$ attention weight for the $j^{th}$ location is:
\begin{align*}
    \pmb{a}^{(n)}_j = 
    \begin{cases}
    \pmb{q}_j \pmb{k}^{(n)T} \ &\text{if } \pmb{m}^{(n)}_j = 1  \\
     -\infty &\text{otherwise}
  \end{cases} \ \ \text{for } j \in \{1, 2, \dots, hw\}.
\end{align*}
and the attention mask $\pmb{m}^{(n)} \in \mathbb{R}^{hw}$ is determined by the $n^{th}$ blob ellipse, where its $j^{th}$ value is 1 if a pixel at location $j$ is within the blob ellipse, and 0 otherwise.

With this masking design, each blob representation and its local visual feature are trained to align with each other, and thus
the model becomes more disentangled. To retain the prior knowledge of pre-trained models for
synthesizing high-quality images, it freezes the weights of the pre-trained diffusion model and only trains the newly added layers.

\begin{figure}[t]
  \centering
    \includegraphics[width=\linewidth]{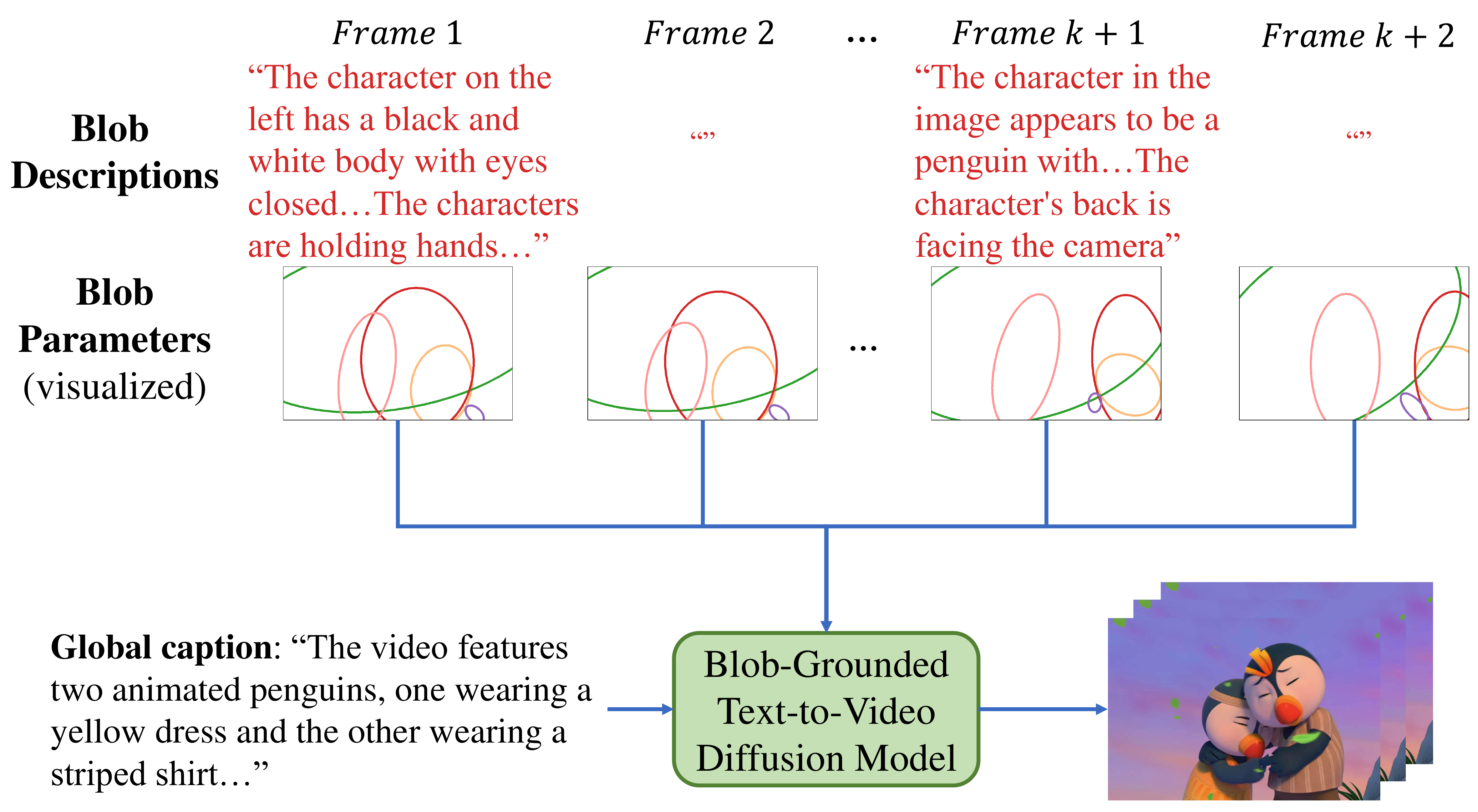}
    \caption{Blob video representations for video generation consist of blob parameters and blob descriptions. Blob parameters exist for every frame while blob descriptions are provided in every $k$ frames. Therefore, only frames $1, k+1,...$ have blob descriptions. }
  \label{fig:blob_repr}
\end{figure}

\begin{figure*}[t]
  \centering
   \includegraphics[width=\linewidth]{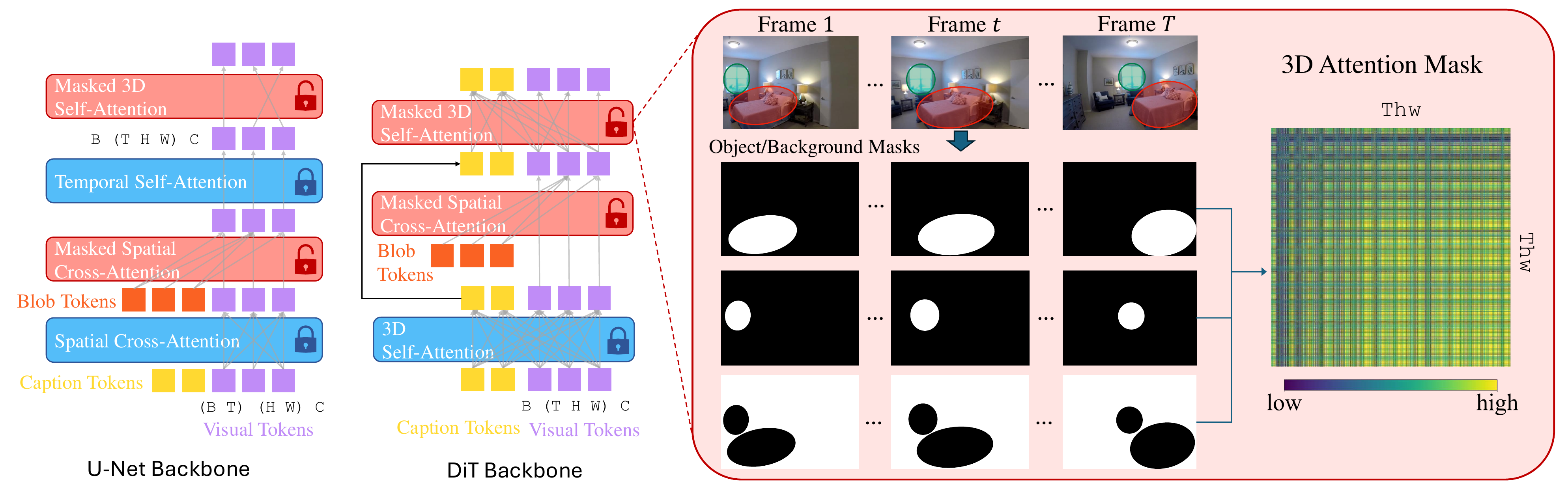}
   \caption{\methodname{} architecture with U-Net backbone or DiT backbone. Our method leverages two masked attention modules that allows: 1) visual features to attend to only corresponding blobs embeddings; 2) the same object attend to itself across frames. High-value elements in the 3D attention mask in the figure will be mapped to 0 while low-value elements are mapped to $-\infty$ as in Eq. \ref{eq:hwt_mask}. Note the multiple colors in the binary 3D attention mask are from the aliasing issue during visualization.
   }
   \label{fig:architecture}
   \vspace{-8pt}
\end{figure*}

\section{Method}\label{sec:method}
We first describe the extension of BlobGEN to video generation, including new blob representations for the video data and new masked spatial cross-attention layers that fuse blob video representations to video diffusion networks. 
Furthermore, we introduce new masked 3D attention layers to improve temporal consistency in the object level. Finally, we present blob video generation based on LLMs, which can serve as a stage before \methodname{} to save human efforts from manually designing layouts.

\subsection{Blob representations for videos}\label{subsec:blob_repre}

Given a video of frame length $T$, we extract objects from the first frame and track each of the extracted objects in subsequent $T-1$ frames. Accordingly, we obtain a \emph{blob video} of the same frame length that contains $N$ blob ellipses in each frame. Similar to BlobGEN, the $n^{th}$ object's spatial features (including shape, size and location) in the $t^{th}$ frame are depicted by blob parameters $\tau^{(n)}_t:= [c_x, c_y, a,b ,\theta]$, defined in the same way as Section~\ref{sec:prelim_blobgen}. 
The blob video captures how the spatial features of each object and their spatial arrangements evolve temporally. On one hand, it can easily capture the object motion in a natural video (e.g., a cat running on the grass), by looking into the relative movement of a blob (e.g., cat) to other blobs (e.g., grass).
On the other hand, it can also capture the camera motion by referring to the joint movements and/or deformations of all blobs.

Similar to BlobGEN, we also pair blobs with free-form text descriptions to provide fine-grained details of the local objects. Compared to previous works that use a single class label for each object across frames~\cite{li2023trackdiffusion, wangboximator}, our blob captions complement the spatial layout with more information such as appearance attributes (color, texture, etc.) and camera focus. 
Besides, since many visual features of an object may change in a video, it becomes very challenging to use a single blob video caption to describe the object appearance and its dynamic variation across frames. Thus, we opt to apply multiple frame-wise object captions for each blob, which are independently extracted from an existing image captioning model. 
However, we do not apply blob captions to every object in every single frame because 1) it is neither efficient in the data annotation stage nor convenient for users to construct during inference, and 2) consecutive frames in most videos have little change in objects' visual features. Instead, we assign blob captions at a fixed interval across time, spacing them every $k$ frames.

In summary, our \emph{blob video representations} in a video are comprised of 1) blob parameters $\{ \tau^{(n)}_t \}$ for every single frame ($t=1,2,\cdots, T$) and every single object ($n=1,2,\cdots, N$), and 2) blob descriptions $\{ s^{(n)}_{t_k} \}$ for every $k$ frame ($t_k = 1,k+1,\cdots, T$) and every single object ($n=1,2,\cdots, N$). 
Particularly, we denote the frames indexed by $t_k$ as \emph{anchor frames} since they contain both blob parameters and blob descriptions. 
As we will show later, we can obtain complete context features for other frames through context interpolations based on the blob captions from anchor frames. This design offers consistent contextual information to avoid modality mismatch while applying our blob video representations.

\subsection{Blob-grounded text-to-video generation}\label{subsec:\methodname{}}
To incorporate blob video grounding into the pre-trained video diffusion models, we follow the design of BlobGEN to add new attention layers to the network. Similarly, we only trained the newly added layers while freezing the weights of pre-trained models. In the following, we introduce the key design choices of \methodname{}.

\paragraph{Context interpolation.} 
To obtain blob embeddings, we follow BlobGEN to encode blob representations for each single frame independently. That is, for the $n^{th}$ object in the $t^{th}$ frame, we first get the blob parameter embedding $\pmb{e}_\tau^{t, n}$ and blob description embedding $\pmb{e}_s^{t, n}$, and concatenate them along the embedding feature dimension as input to an MLP network for its blob embedding $\pmb{e}_{\text{blob}}^{t, n}$. 
However, not all frames are paired with blob captions, which means we do not have blob description embedding $\pmb{e}_s^{t, n}$ for those non-anchor frames whose frame index $t \neq t_k$. 

A naive approach is to encode an empty text string with CLIP text encoder and use it as the blob description embedding for all non-anchor frames. But it can easily introduce inconsistency across frames due to the large contextual mismatch. To overcome this issue, we propose a simple method called \emph{context interpolation} that linearly interpolates the blob description embeddings of two consecutive anchor frames for each non-anchor frame in the middle. Formally, given the indices of two anchor frames $t_k$ and $t_{k+1}$ where $t_{k+1} = t_k + k$, the interpolated blob description embedding of the non-anchor frame indexed by $t \in (t_{k}, t_{k+1})$ is given by 
\begin{align}\label{eq:linear_interp}
    \pmb{e}_s^{t, n} = \frac{t_{k+1}-t}{k} \pmb{e}_s^{t_{k+1}, n} + \frac{t - t_{k}}{k} \pmb{e}_s^{t_{k}, n}
\end{align}
Intuitively, this linear interpolation ensures a smooth semantic transitioning of object captions across all frames in the CLIP embedding space, leading to better temporal consistency and blob-guided controllability. Besides linear interpolation, some learnable nonlinear interpolations can also be considered. For example, we can train a Perceiver IO network \cite{jaegle2021perceiver} that takes the blob description embeddings of anchor frames as input and learns the blob descriptions embeddings of other frames.

\paragraph{Masked spatial cross-attention.} 
The extension of masked cross-attention from BlobGEN to fuse blob video representations with video features is straightforward. Similar to spatial attention layers in many video diffusion backbones~\cite{chen2024videocrafter2}, both the visual features and blob embeddings are first reshaped in the form of $(\texttt{B T (h w) c}) \rightarrow (\texttt{(B T) (h w) c})$ and then they can be fused by applying the masked cross-attention in Eq. (\ref{maskca}). That is, we fuse blob embeddings and visual features in the same frame independently for all the frames. This design makes the masked spatial cross-attention layers to solely focus on promoting the frame-wise alignment of generated content and the blob conditioning, without worrying about temporal consistency.

\paragraph{Masked 3D self-attention.} The masked spatial cross-attention can only apply per-frame consistency between frames and blobs and cannot guarantee temporal consistency across frames. To improve temporal consistency, we propose new masked 3D self-attention layers to enforce object-level temporal consistency. 
Note that even though many video diffusion models~\cite{blattmann2023stable} based on U-Net are equipped with temporal self-attention, it only allows each ``pixel'' of the visual feature map in a frame to attend to ``pixels'' at the same spatial location in other frames. {However, blobs provide a rough location of each object over time, and thus we can impose stronger coherence by biasing the attention towards the same object over time.}

Specifically, in masked 3D self-attention, we flatten all three dimensions in a video feature (i.e., $T, h, w$) into one dimension and denote the resulting feature as $\pmb{g}\in\mathbb{R}^{Thw\times d}$. Then we obtain query, key and value with three linear projections for self-attention as $\pmb{q}=\pmb{g}\pmb{W}_q$, $\pmb{k}=\pmb{g}\pmb{W}_k$, $\pmb{v}=\pmb{g}\pmb{W}_v$, all in the shape of $\mathbb{R}^{Thw\times d}$. 
Then, the masked 3D self-attention can be written as:
\begin{equation}\label{eq:mask_3d}
    \text{MaskSA3D} := \text{Softmax}(\frac{\pmb{q}\pmb{k}^T}{\sqrt{d}}+\pmb{M}_{\text{blob}})\pmb{v},
\end{equation}
where $\pmb{M}_{\text{blob}}\in\mathbb{R}^{Thw\times Thw}$ is a 3D mask determined by blob ellipses across frames, which we describe in the next. 

Similar to BlobGEN, we denote the binary blob mask for the $n^{th}$ object in the $t^{th}$ frame as $\pmb{m}^{t,n} \in \mathbb{R}^{hw}$, where its $i^{th}$ entry (denoted as $\pmb{m}^{t,n}_i$) equals 1 if the location $i$ is within the blob ellipse, and 0 otherwise. Besides the $N$ blob masks corresponding to $N$ objects in each frame, we introduce another binary mask, called \emph{background mask}, as $\pmb{m}^{t, \text{bg}}=1-\bigcup_{n=1}^{N}\pmb{m}^{t, n}$, resulting in $N+1$ blob masks that cover the whole ($h \times w$) spatial space. Given any two indices $i,j \in \{1, 2, \cdots, Thw\}$, we then define each entry of $\pmb{M}_{\text{blob}}$ indexed by $(i, j)$ as

\begin{equation}\label{eq:hwt_mask}
    \pmb{M}_{\text{blob}}^{i,j}=
     \begin{cases} 
      0 & \text{if} \;\;  \pmb{m}_{i}^{t,n} \land  \pmb{m}_{j}^{t',n}=1, \;\; \forall t, t',n \\
      0 & \text{if} \;\;  \pmb{m}_{i}^{t,\text{bg}} \land \pmb{m}_{j}^{t',\text{bg}}=1, \;\; \forall t, t' \\
      -\infty & \text{otherwise}
    \end{cases}
\end{equation} 
which allows the local object feature for the $t$ frame (depicted by a blob ellipse) to only attend to local features of the same object for another frame (including the  $t$ frame itself). Note that each background feature 
only attends to other background features across frames.
Thus, this 3D mask design implies an object-centric self-attention mechanism, leading to better object-level cross-frame consistency. Furthermore,
the use of $\pmb{m}^{t,\text{bg}}$ is critical in practical implementation to avoid having all-zero rows in the input to the softmax function and improve training stability.

Fig. \ref{fig:architecture} shows the overview architecture of \methodname{}. The introduced two types of attention modules can be inserted into both U-Net and DiT-based diffusion models with minimal modification. We always arrange our masked 3D self-attention after the masked spatial cross-attention as a bottleneck for context feature fusion.

\subsection{LLMs for blob generation} 

Inspired by previous work in using LLMs for layout planning \cite{lianlmd, lianllm, lin2023videodirectorgpt}, we also generate video layouts with in-context learning and structured text. Since video layouts need to expand over time dimension and may have multiple objects per frame, it is important to find a robust structure to represent them. Instead of using self-defined template \cite{tian2024videotetris} or stylesheet language \cite{feng2024layoutgpt}, we form the layouts as nested dictionaries where frame index, object id, blob parameters and captions are settled in different layers of the structure. LLMs interpret and generate outputs in the same json format that can be directly parsed into blob layouts per frame. In addition, we only generate blobs for a sparse set of frames while interpolate the intermediate blob parameters to make the stage more efficient. We append our detailed in-context prompts in the Appendix. 
\section{Experiment} \label{sec:experiment}
\subsection{Experiment setup}
\paragraph{Data preparation. } Since there are no available video datasets that provides ground truth blob annotations align with our setting in Sec.~\ref{subsec:blob_repre}, we build an annotation pipeline to extract blob parameters and captions. In general, we apply Grounding DINO \cite{liu2023grounding} or ODISE \cite{xu2023open} to the first frame of each video and obtain segmentation masks. Then we apply SAM2 \cite{ravi2024sam} to every other frame to track the objects. After obtaining all segmentation masks throughout the video, we fit an ellipse to every mask by optimizing the Intersection Over Union (IOU) between the ellipse and the mask area. Using the segmentation masks, we also crop out objects in every eight frames and apply LLaVA-NeXT \cite{liu2024llavanext} to get blob captions in these frames. 

Our video-text pairs mainly come from OpenVid-1M~\cite{nan2024openvid}, VidGEN-1M~\cite{tan2024vidgen}, and a small subset of HD-VILA~\cite{xue2022advancing}. We apply heavy filtering to each stage of the annotation pipeline to maintain high quality and good balance between human and non-human objects. We end up with $\sim$1M videos having dense blob annotations for training. For indoor scene videos, we use the processed ScanNet++~\cite{yeshwanth2023scannet++} from BlobGEN-3D \cite{Liu2024BlobGen3D} where the frame blobs are projected from 3D blobs fitted on the point cloud segmentation extracted from scenes.

\begin{figure*}
    \centering
    \includegraphics[width=\linewidth]{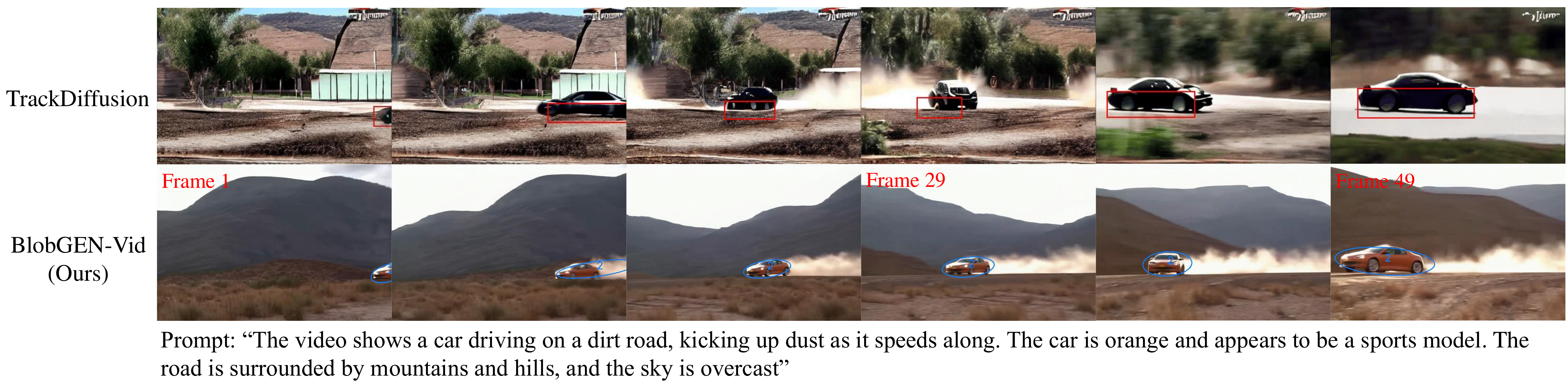}
    \caption{Layout-to-video generation results on YoutubeVIS-2021~\cite{vis2021}. The visualized layouts are ground truth layouts fed into the models during inference. Our method shows better prompt-video alignment than the strongest baseline TrackDiffusion~\citep{li2023trackdiffusion}. }
    \label{fig:layout-to-video}
\end{figure*}

\paragraph{Benchmarks and metrics. } We have three different experiment settings/domains. First, we test \textbf{layout-to-video} generation on 717 validation and test videos of Youtube-VIS 2021~\cite{vis2021}. We report FVD \cite{unterthiner2018towards} against the ground truth 717 videos for general visual quality, mean Intersection-over-Union (mIOU) for layout controllability, rCLIP$_t$ and rCLIP$_i$ for prompt-video alignment, and regional cross-frame CLIP similarity (rCFC) for object consistency. rCLIP$_t$ is the CLIP cosine similarity between each blob caption and blob-bounded region in the generated videos, and rCLIP$_i$ is the similarity between regions from generated videos and ground truth videos. Secondly, to evaluate \textbf{compositionality in T2V} setting, we report results on T2V-CompBench~\cite{sun2024t2v} and TC-Bench~\cite{feng2024tc}, which evaluate composition changes over time in different aspects. They both equip large multimodal models or detection/tracking models for different aspects. Lastly, we evaluate \textbf{multi-view indoor} videos on 1392 test videos from ScanNet++. We report FID and IS for frame quality, FVD for video quality, PSNR and cross-frame CLIP similarity (CFC) and rCFC for consistency. For details of our data and evaluation setup, refer to the Appendix.

\subsection{Layout-grounded video generation. } 

\begin{table}[t]
\centering
\resizebox{\linewidth}{!}{%
\begin{tabular}{l ccccc}
\toprule
    \multirow{2}{*}{\textbf{Method}} &\multicolumn{5}{c}{\textbf{YoutubeVIS-2021}\cite{vis2021}} \\
    \cmidrule{2-6}
    & FVD $\downarrow$ & mIOU $\uparrow$ & rCLIP$_t$ $\uparrow$ & rCLIP$_i$ $\uparrow$ & rCFC $\uparrow$ \\
    \midrule
    TrackDiffusion \cite{li2023trackdiffusion} & 464 & 0.4916 & 0.2731 & 0.8041 & 0.9403 \\
    LVD \cite{lianllm} & 558 & 0.2814 & 0.2613 & 0.8028 & 0.8608 \\
    VideoTetris \cite{tian2024videotetris} & 590 & 0.1658 & 0.2669 & 0.7991 & 0.4231 \\
    \midrule
    \methodname{} (VC2) & 396 & \textbf{0.6119} & 0.2794 & 0.8223 & 0.9491 \\
    \methodname{} (CogVideoX-5B) & \textbf{317} & 0.5982 & \textbf{0.2888} & \textbf{0.8364} & \textbf{0.9580} \\
    \bottomrule
\end{tabular}
}
\caption{Evaluation results of layout-guided video diffusion models on YoutubeVIS-2021 benchmarks.}
\label{tab:layout2video_results}
\end{table}

As is shown in Table~\ref{tab:layout2video_results}, our method outperforms all baselines in nearly all aspects of evaluation. In particular, our method based on VC2 achieves the highest mIOU score (0.6119), over 20\% improvement in spatial controllability over TrackDiffusion \cite{li2023trackdiffusion}. \methodname{} based on CogVideoX-5B achieves a slightly lower mIOU score as it is trained on only half of the dataset but still outperforms all baselines. As for prompt-video alignment, our method outperforms all baselines by achieving 0.2888 rCLIP$_t$ and 0.8364 rCLIP$_i$ scores. As our blob captions are free-form language descriptions of the objects instead of a coarse-grained category name or id, they provide rich semantics to facilitate control of fine-grained details.

We show an qualitative comparison between TrackDiffusion (TD) and our method in Fig. \ref{fig:layout-to-video}. \methodname{} tightly follow the blobs to generate the car while the objects in TD's video often reach out of the box. In addition, our method also demonstrate better prompt-video alignment by showing ``orange sport car'' while the baseline cannot control such semantics because it uses the category label ``car''. 

\subsection{Text-to-video generation}

\begin{table*}[t]
\centering
\resizebox{\textwidth}{!}{%
\begin{tabular}{l c ccccc c cccccc}
\toprule
    \multirow{3}{*}{\textbf{Method}} && \multicolumn{5}{c}{\textbf{T2V-CompBench \cite{sun2024t2v}}} && \multicolumn{6}{c}{\textbf{TC-Bench \cite{feng2024tc}}} \\
    \cmidrule{3-7} \cmidrule{9-14}
    && \multirow{2}{*}{Consist-Attr $\uparrow$} & \multirow{2}{*}{Dynamic-Attr $\uparrow$} & \multirow{2}{*}{Spatial $\uparrow$} & \multirow{2}{*}{Motion $\uparrow$} & \multirow{2}{*}{Numeracy $\uparrow$} && \multicolumn{2}{c}{Attr. Transition} & \multicolumn{2}{c}{Obj. Relation} & \multicolumn{2}{c}{Background Shift} \\
    \cmidrule{9-10}\cmidrule{11-12}\cmidrule{13-14}
    &&&&&&&& TCR $\uparrow$ & TC-Score $\uparrow$ & TCR $\uparrow$ & TC-Score $\uparrow$ & TCR $\uparrow$ & TC-Score $\uparrow$ \\
    \midrule
    Open-Sora v1.2 \cite{opensora} && 0.6600 & 0.1714 & 0.5406 & 0.2388 & 0.2556 && 6.15 & 0.6509 & 7.66 & 0.7406 & 2.35 & 0.5847 \\
    CogVideoX-5B \cite{yang2024cogvideox} && - & - & - & - & - && 8.08 & 0.6930 & 10.64 & 0.7237 & 4.71 & 0.6338 \\
    LVD~\cite{lianllm} (w/ GPT-4) && 0.5595 & 0.1499 & 0.5469 & 0.2699 & 0.0991 && 5.77 & 0.6215 & \textbf{12.77} & 0.7081 & 1.96 & 0.5042 \\ 
    VideoTetris~\cite{tian2024videotetris} (w/ LLM) && 0.7125 & 0.2066 & 0.5148 & 0.2204 & 0.2609 && - & - & - & - & - & - \\
    \midrule
    Pika 1.0 && 0.6513 & 0.1744 & 0.5043 & 0.2221 & 0.2613 && 5.77 & 0.6520 & 8.51 & 0.7242 & 1.96 & 0.6070 \\
    Dream Machine && 0.6900 & 0.2002 & 0.5387 & 0.2713 & 0.2109 && 9.80 & 0.7319 & \textbf{12.77} & 0.7755 & 5.88 & 0.6284  \\
    Kling 1.0 && \textbf{0.8045} & 0.2256 & 0.6150 & 0.2448 & 0.3044 && 7.69 & 0.6888 & 10.64 & 0.7819 & 3.92 & 0.6183 \\
    Gen-3 Alpha && 0.7045 & 0.2078 & 0.5533 & 0.3111 & 0.2169 && 9.62 & \textbf{0.7507} & 10.64 & 0.7073 & \textbf{27.45} & \textbf{0.7488} \\
    GPT-4o + \methodname{} (Ours) && 0.7400 & \textbf{0.2650} & \textbf{0.6725} & \textbf{0.3880} & \textbf{0.3910} && \textbf{15.39} & 0.7055 & \textbf{12.77} & \textbf{0.7944} & 10.42 & 0.6852 \\
    \bottomrule
    
\end{tabular}
}
\caption{Comparison between \methodname{} and major proprietary text-to-video diffusion models/systems on two benchmarks emphasizing video compositionality: T2V-CompBench~\citep{sun2024t2v} and TC-Bench~\citep{feng2024tc}.}
\label{tab:text2video_results}
\vspace{-12pt}
\end{table*}

Table \ref{tab:text2video_results} shows the evaluation results of our complete text-to-video generation pipeline by combining GPT-4o and \methodname{} (CogVideoX-5B-based). We adopt in-context learning methods and input two fixed exemplars to GPT-4o to obtain blob parameters and blob captions for a sparse set of frames. Then we linearly interpolate blob parameters and feed the blob conditions into \methodname{} to generate videos. Our pipeline outperforms proprietary video generators in four challenging compositional issues, including dynamic attribute binding, spatial relation accuracy, motion binding and numerical accuracy. We show more qualitative examples in the Appendix.

\subsection{Multi-view scene generation}

\begin{table}
  \centering
  \resizebox{\linewidth}{!}{%
  \begin{tabular}{ll cccc}
    \toprule
    & \multicolumn{1}{c}{\multirow{2}{*}{\textbf{Method}}} && \multicolumn{3}{c}{\textbf{Image-based Metrics}} \\
    \cmidrule{4-6}
    &&& FID $\downarrow$ & IS $\uparrow$ & CLIP Sim. $\uparrow$ \\
    \midrule
    \texttt{1} & BlobGEN-3D (blob only) && \textbf{21.24} & 5.38 & 0.2301 \\
    \texttt{2} & BlobGEN-3D \cite{Liu2024BlobGen3D} && 31.28 & 5.02 & 0.2231 \\
    \texttt{3} & \methodname{} w/o Mask 3D Attn && 30.72 & 5.66 & 0.2319 \\
    \texttt{4} & \methodname{} (Ours) && 27.94 & \textbf{5.70} & \textbf{0.2320} \\
    \midrule
    \midrule
    && \multicolumn{4}{c}{\textbf{Video-based Metrics}} \\
    \cmidrule{3-6}
    && FVD $\downarrow$ & PSNR $\uparrow$ & CFC $\uparrow$ & rCFC $\uparrow$ \\
    \midrule
    \texttt{5} & BlobGEN-3D (blob only) & 335 & 10.06 & 0.9168 & 0.9197 \\
    \texttt{6} & BlobGEN-3D \cite{Liu2024BlobGen3D} & 468 & 15.23 & 0.9322 & 0.9347 \\
    \texttt{7} & \methodname{} w/o Mask 3D Attn & \textbf{142} & 21.71 & 0.9432 & 0.9424 \\
    \texttt{8} & \methodname{} (Ours) & 161 & \textbf{22.20} & \textbf{0.9453} & \textbf{0.9456} \\
    \bottomrule
  \end{tabular}
  }
  \caption{Evaluation results on ScanNet++ \cite{yeshwanth2023scannet++} test split with image and video metrics. \methodname{} is based on VC2.}
  \label{tab:indoor_results}
  \vspace{-10pt}
\end{table}

For multi-view indoor scene generation, we directly compare to BlobGEN-3D~\cite{Liu2024BlobGen3D} as shown in Table~\ref{tab:indoor_results}. BlobGEN-3D is fine-tuned from BlobGEN~\cite{niecompositional} with a depth-conditioned ControlNet~\cite{zhang2023adding} and a warped previous frame. It is an image diffusion model that generates free-view indoor images in an autoregressive frame-by-frame manner. 

We can see \methodname{} outperforms BlobGEN-3D in all metrics, especially in video consistency. While BlobGEN-3D without depth condition (row 2 and 5) achieves the lowest FID score, it fails to maintain cross-frame consistency as indicated by the low PSNR (10.06) and CFC values (0.9168). When our proposed masked 3D attention is removed from \methodname{} (comparing row 7 and 8), PSNR and CFC both decrease, justifying the effectiveness of 3D masks in improving consistency.   

\begin{figure}[t]
  \centering
   \includegraphics[width=\linewidth]{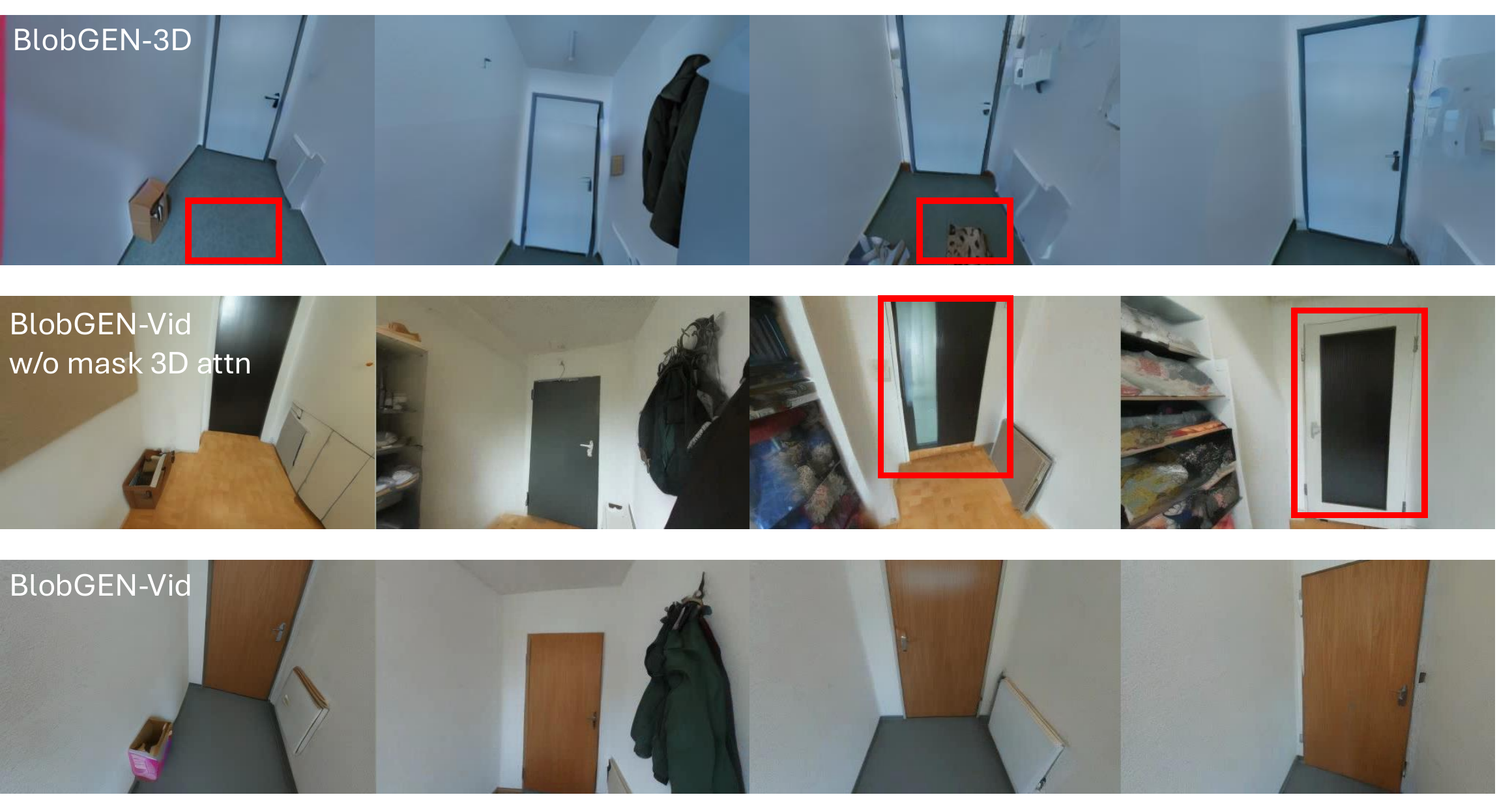}
   \caption{Qualitative results on ScanNet++, where our method, especially with masked 3D attention, shows much better consistency in the door appearance than BlobGEN-3D.}
   \label{fig:scannetpp_qualitative}
   \vspace{-10pt}
\end{figure}

We also show qualitative comparison in Fig.~\ref{fig:scannetpp_qualitative}. The red boxes annotate inconsistent objects in the generated image sequences. \methodname{} without masked 3D self-attention tend to generate the door in different colors as the camera pose changes, while \methodname{} generate more consistent appearance of the door.

\subsection{Ablation study}

\begin{table}
  \centering
  \resizebox{\linewidth}{!}{%
  \begin{tabular}{l ccc| ccccc}
    \toprule
    & \shortstack{Mask \\3D Attn} & \shortstack{Context\\Interp.} & \shortstack{Training\\Data} & FVD $\downarrow$ & mIOU $\uparrow$ & rCLIP$_t$ $\uparrow$ & rCLIP$_i$ $\uparrow$ & rCFC $\uparrow$ \\
    \midrule
    \texttt{1} && Linear & 160K & 617 & 0.2585 & 0.2697 & 0.7961 & 0.9232 \\
    \texttt{2} & $\checkmark$ & Linear & 160K & 368 & 0.5623 & 0.2804 & 0.8211 & \textbf{0.9500} \\
    \midrule
    \texttt{3} & $\checkmark$ & Linear & 400K & \textbf{346} & 0.5771 & 0.2794 & 0.8200 & 0.9480 \\
    \texttt{4} & $\checkmark$ & Slerp & 400K  & 378 & 0.5702 & 0.2763 & 0.8142 & 0.9438 \\
    \texttt{5} & $\checkmark$ & Perceiver & 400K & 352 & 0.5926 & \textbf{0.2806} & 0.8204 & 0.9459 \\
    \midrule
    \texttt{6} & $\checkmark$ & Perceiver & 1M & 396 & \textbf{0.6119} & 0.2794 & \textbf{0.8223} & 0.9491\\
    \bottomrule
  \end{tabular}
  }
  \caption{Ablation study on model architecture, interpolation method and training data size. The base model is VideoCrafter2. 
  }
  \label{tab:ablation}
  \vspace{-10pt}
\end{table}

In Table~\ref{tab:ablation}, we present the ablation study on three factors: Masked 3D self-attention, context interpolation method, and training data. We observe that adding the masked 3D self-attention is crucial to facilitate video diffusion models to generate consistent objects, as indicated by the significant discrepancy between rows 1 and 2 in all metrics. Specifically, adding masked 3D attention improves FVD by 40\% and mIOU by 117\% with the same amount of data and interpolation method. 

As for context interpolation, we investigate three different approaches including linear, slerp and PerceiverIO. While slerp has been widely used in GANs to interpolate latent features \cite{karras2019style}, we do not find any advantage of it in our setting. We conjecture that the embedding space of CLIP text encoder may have a different topology that makes it less effective. Linear and perceiver-based interpolation illustrate slightly different behavior. The former achieves a larger rCFC value while PerceiverIO shows stronger layout controllability as indicated by the mIOU values. 

Finally, we do observe the effectiveness of scaling training data from 160K to 400K and eventually to 1M videos. The effects are mainly reflected as stronger layout controllability as indicated by mIOU values. Models trained with 400K videos (row 3-5) all show better mIOU compared to row 1-2. Increasing the data to 1M (row 6) further boosts mIOU by 3.2\% and rCLIP$_i$ and rCFC as well. It is important to realize that FVD may not be a robust metric \cite{ge2024content} and the fact that higher rCFC values do not necessarily indicate better performance~\cite{feng2024tc}. 

\vspace{-5pt}
\section{Conclusions}\label{sec:conclusion}
In this work, we propose a new layout design for text-to-video generation, called blob representations. The representation contains blob parameters for each object in every frame and paired blob captions in a sparse set of frame. 
Our free-form blob captions also provide more fine-grained semantics of each object. We then introduce a framework termed \methodname{} that endows video diffusion models with the ability to condition on blob inputs. \methodname{} consists of a context interpolation module, allowing more flexible semantic transition, and masked 3D attention blocks to enforce object consistency across frames. 
We demonstrate the effectiveness of our frame in multiple visual domains and settings. \methodname{} achieves strong performance in open-domain video generation and multi-view image generation. When combined with an LLM, it shows great potential in compositionality and outperforms proprietary video generators in multiple aspects. 

{
    \small
    \bibliographystyle{ieeenat_fullname}
    \bibliography{main}
}

\maketitlesupplementary
\appendix

\newpage
\section{Data annotation}\label{appendix:dataset}

\begin{figure}[t]
  \centering
    \includegraphics[width=\linewidth]{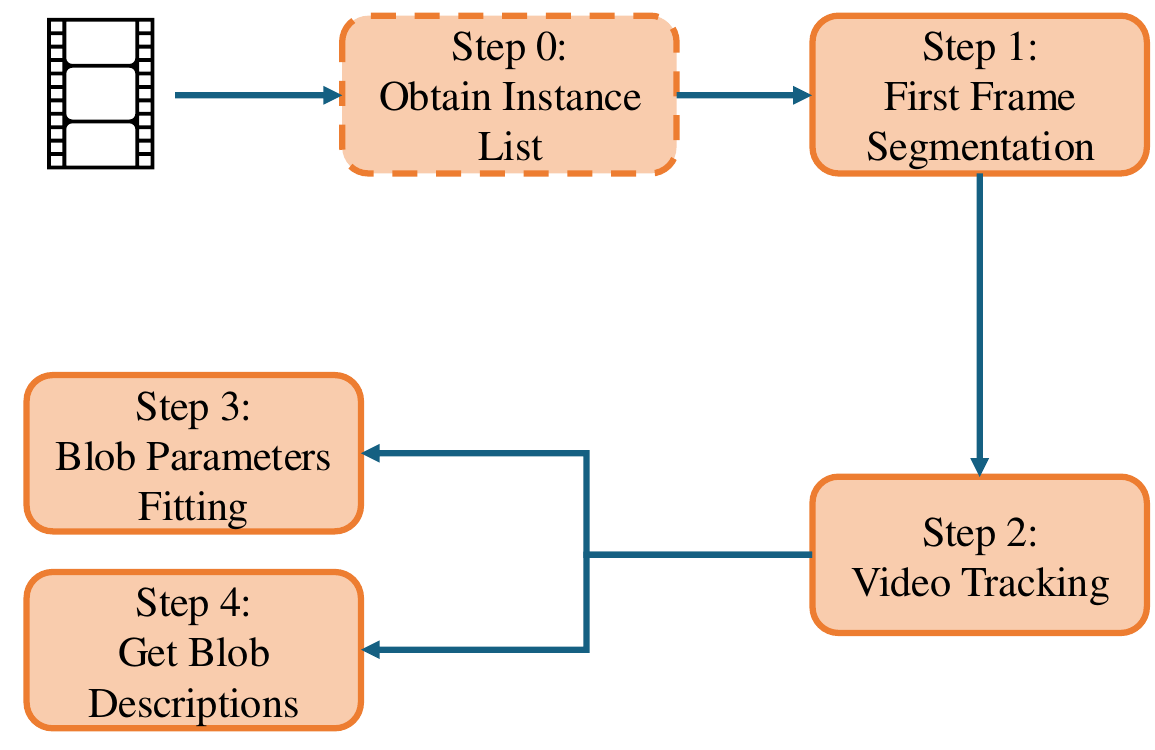}
    \caption{Our data annotation pipeline for obtaining blob video representations consists of five steps with step 0 being optional.}
  \label{fig:appendix_data}
\end{figure}

\paragraph{Data annotation pipeline. } Our data annotation pipeline consists of four to five key steps as shown in Fig. \ref{fig:appendix_data}. The step 0 is to obtain a list of objects appeared in the video using a VLM. Though previous methods~\cite{wangboximator} directly use a language parser to extract object nouns or phrases from video captions, we found the parser often extracts words that do not represent concrete entities and introduces additional noise to step 1. Thus we feed videos into LLaVA-NeXT-Video-7B~\cite{zhang2024llavanext-video} and prompt it to generate a list of objects that appear in each video. Using the instant list, we could apply Grounding DINO \cite{liu2023grounding} in step 1 to obtain segmentation masks for the first frame of the video. We also experiment with ODISE~\cite{xu2023open}, which is a panoptic segmentation model that does not require the instance list from step 0 to work. For most videos, we first apply LLaVA-NeXT+Grounding DINO to get segmentation masks. If the mask coverage is below 20\% of the frame size, we apply ODISE to get more dense panoptic annotation. This helps us keep most of the videos for further annotation and hence improve data utilization rate. 

After obtaining the segmentation masks in step 1, we apply SAM2~\cite{ravi2024sam} to track each object mask throughout the video. To make the process efficient, we uniformly sample 1/4 of all the frames to do tracking. With the tracking masks for the frames, we can fit a set of blob parameters ($c_x, c_y, a,b,\theta$) for each mask. For frames without tracking masks, we linearly interpolate the blob parameters from the closest neighboring frames. As for step 4, we crop a tight rectangle region around each segmentation mask, and feed it to LLaVA-v1.6-mistral-7b~\cite{liu2024llavanext} to get blob descriptions. For efficiency, we only annotate blob descriptions for the first of every eight frames.  

\begin{figure*}[t]
  \centering
    \includegraphics[width=\textwidth]{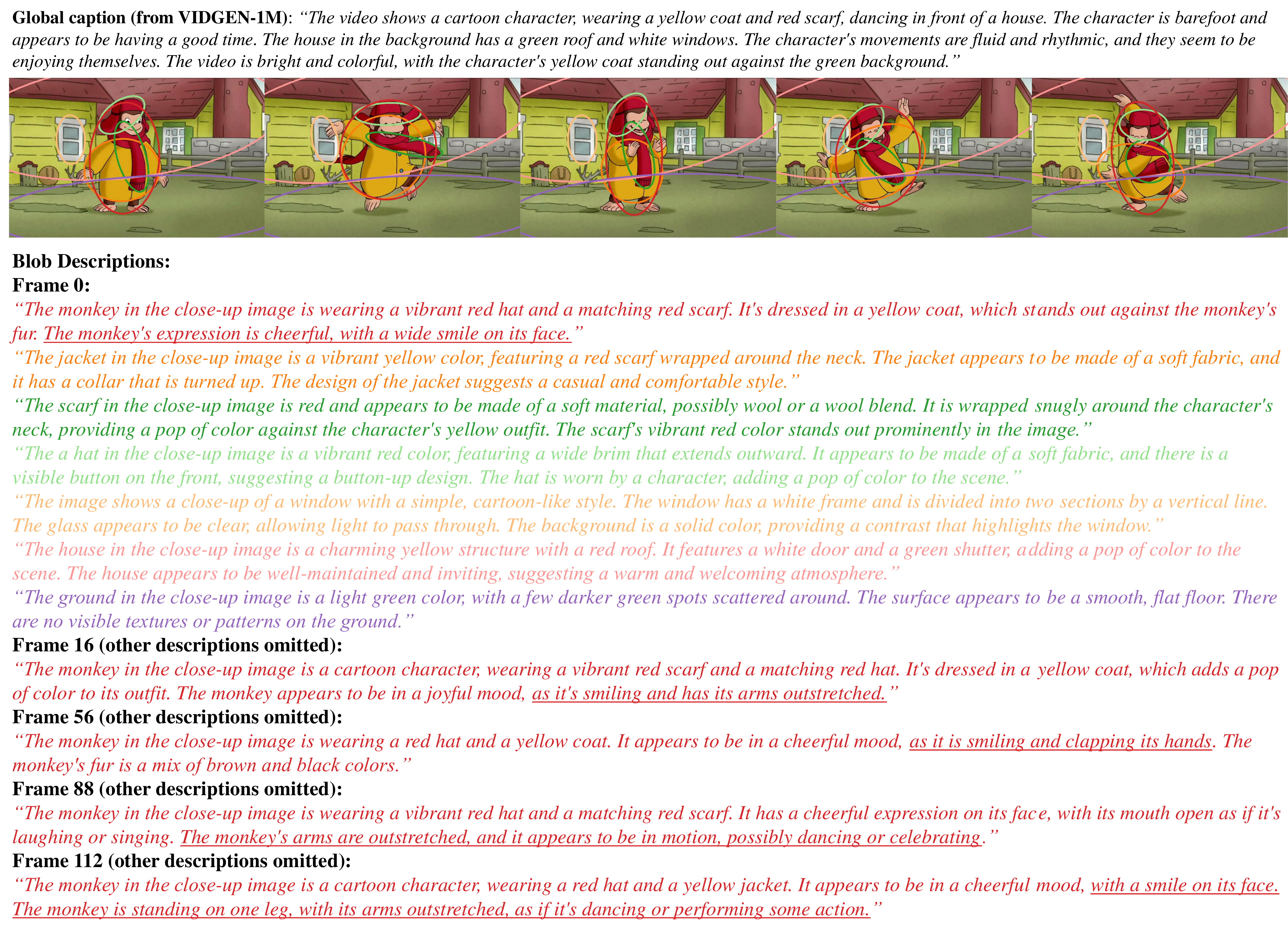}
    \caption{An example of our data annotation results using instance list and Grounding DINO for segmentation. The text color of the blob descriptions match with the blob colors in the frames. The underlined text highlights the changing part of the descriptions as the monkey's gesture and expression changes over time.}
  \label{fig:appendix_data_example1}
\end{figure*}

\begin{figure*}[t]
  \centering
    \includegraphics[width=\textwidth]{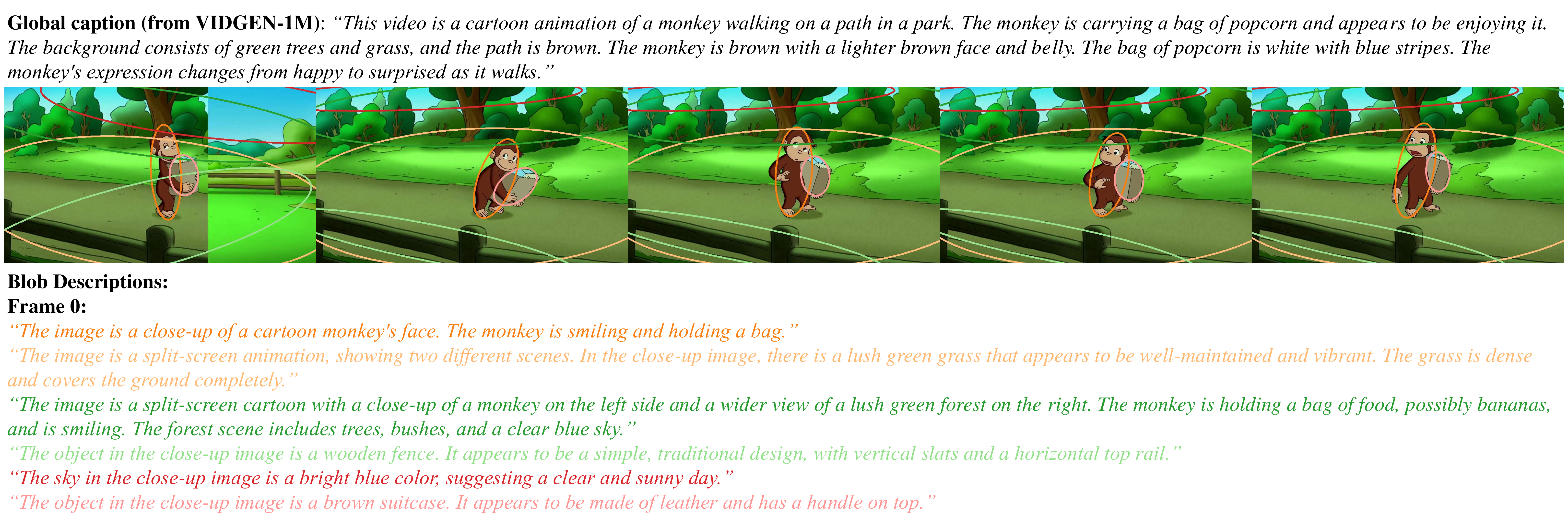}
    \caption{An example of our data annotation results using ODISE as the panoptic segmentation model. ODISE tends to segment the background into different parts, including the sky, trees, grass, and fence in this example.}
  \label{fig:appendix_data_example2}
\end{figure*}

\paragraph{Qualitative visualization. }
We visualize two example of our data annotation results in Fig.~\ref{fig:appendix_data_example1} (using Grounding DINO) and Fig.~\ref{fig:appendix_data_example2} (using ODISE). We observe that the instance list obtained from LLaVA-NeXT-Video-7B~\cite{zhang2024llavanext-video} usually contain instance names in different hierarchical levels. For example, in Fig.~\ref{fig:appendix_data_example2}, the hat, scarf, and yellow jacket are listed as separate objects. Sometimes, the model would also list ``hands'' as separate objects from the whole human figure. However, it has the drawback of neglecting background objects even though we explicitly emphasize ``both foreground and background'' objects in the prompt. 

In contrast, ODISE~\cite{xu2023open} has more fine-grained segmentation of background since it applies a long list of category names merged from different datasets. As is shown in Fig.~\ref{fig:appendix_data_example2}, ODISE segments the background into four different parts, including the sky, trees, grass and fence. However, ODISE's category set does not include some general objects or hierarchical parts of objects like ``cartoon character'' or ``hands/arms'' compared to using instance list. In addition, the segmentation labels from ODISE can be less accurate. For example, it annotates the ``cartoon monkey'' as ``costume'' and the ``brown bag'' as ``suitcase''. The issue is mitigated as we use an VLM to obtain free-form blob descriptions instead of adopting ODISE labels.

\section{BlobGEN-Vid framework}\label{appendix:method}
\begin{figure*}[t]
  \centering
    \includegraphics[width=\linewidth]{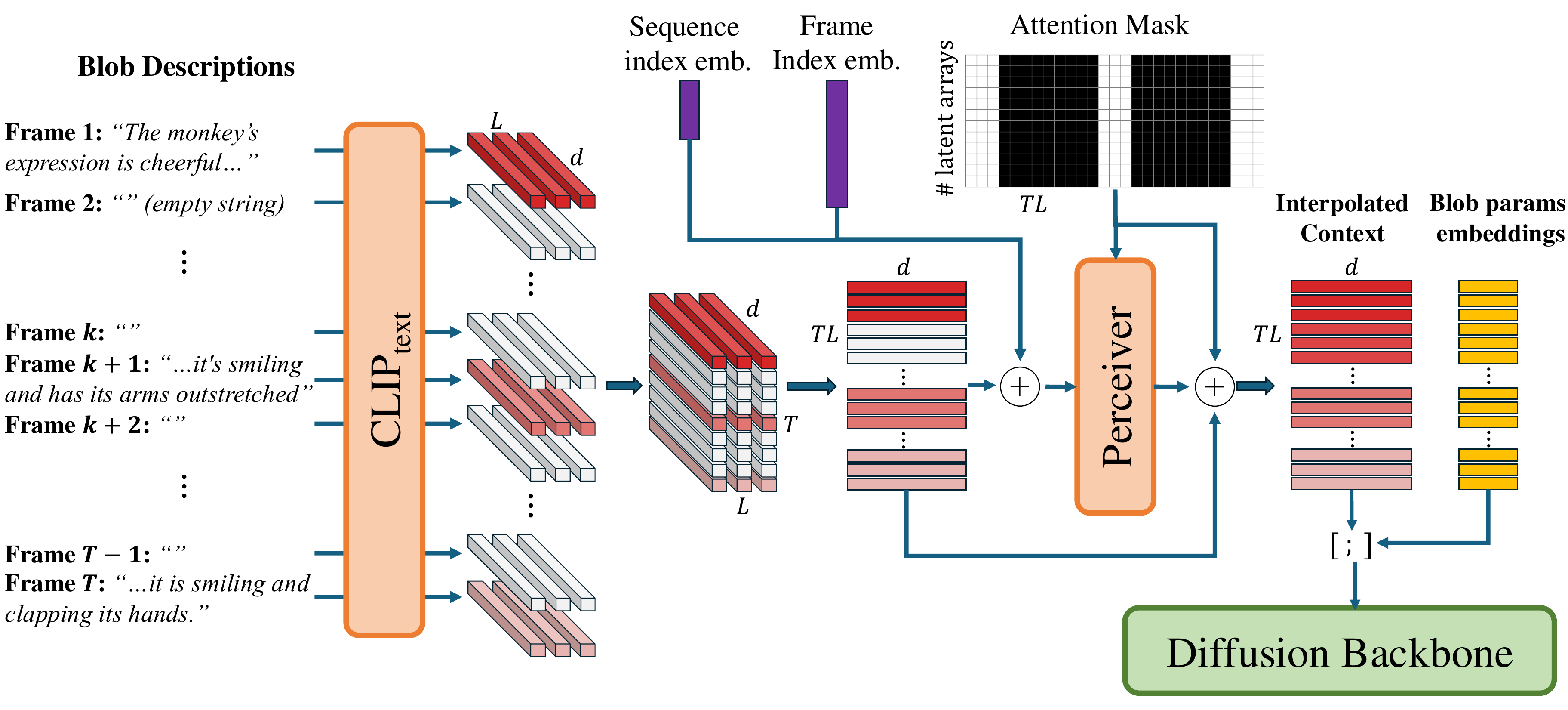}
    \caption{An illustration of the context interpolation stage using a Perceiver-based model. Note that we omit the batch size ($B$) and number of blobs per frame dimension ($N$) for simplicity. After the CLIP text encoder encodes each blob description, we merge time $T$ and context sequence length $L$ into one dimension and learn the context for non-anchor frames through the Perceiver module. The attention mask prevents the latent arrays to attend on blob descriptions that are empty strings, implying that Perceiver only relies on anchor frames' text embeddings to infer intermediate text embeddings.}
  \label{fig:appendix_interp}
\end{figure*}

\paragraph{Context interpolation module. } As shown in Fig.~\ref{fig:appendix_interp}, we first encode the blob descriptions in all frames. For non-anchor frames, we use empty strings for the encoding process and later replace them with the learned features. If an object undergoes apparent semantic change (e.g. object changing color), the blob description in the anchor frames would have different meaning, reflected as the color differences in the penultimate feature sequence from CLIP text encoder. Apart from the linear interpolation introduced in Sec. 4.2, we also experiment with a learnable module using PerceiverIO~\cite{jaegle2021perceiver}. To ensure object-wise interpolation, we reshape the context embeddings as $\texttt{(B T N L d)}\rightarrow \texttt{((B N) (T L) d)}$ where $T$ denotes the number of latent frames and $L$ is the sequence length of the context features. In Fig.~\ref{fig:appendix_interp}, we have omitted $B, N$ and use $L=3$ and $T=9$ for demonstration purpose. In our implementation, the CLIP text encoder outputs $L=77$ context features and there are $T=13$ (CogVideoX) or $T=16$ (VC2) frames. While PerceiverIO was originally proposed to handle inputs of different modalities, we adopt it for the sake of simplicity and flexibility, as it allows arbitrary number of anchor frames. It facilitates handling arbitrary number and locations of the anchor frames on users' choices in the future.

\begin{figure*}[t]
  \centering
    \includegraphics[width=0.9\linewidth]{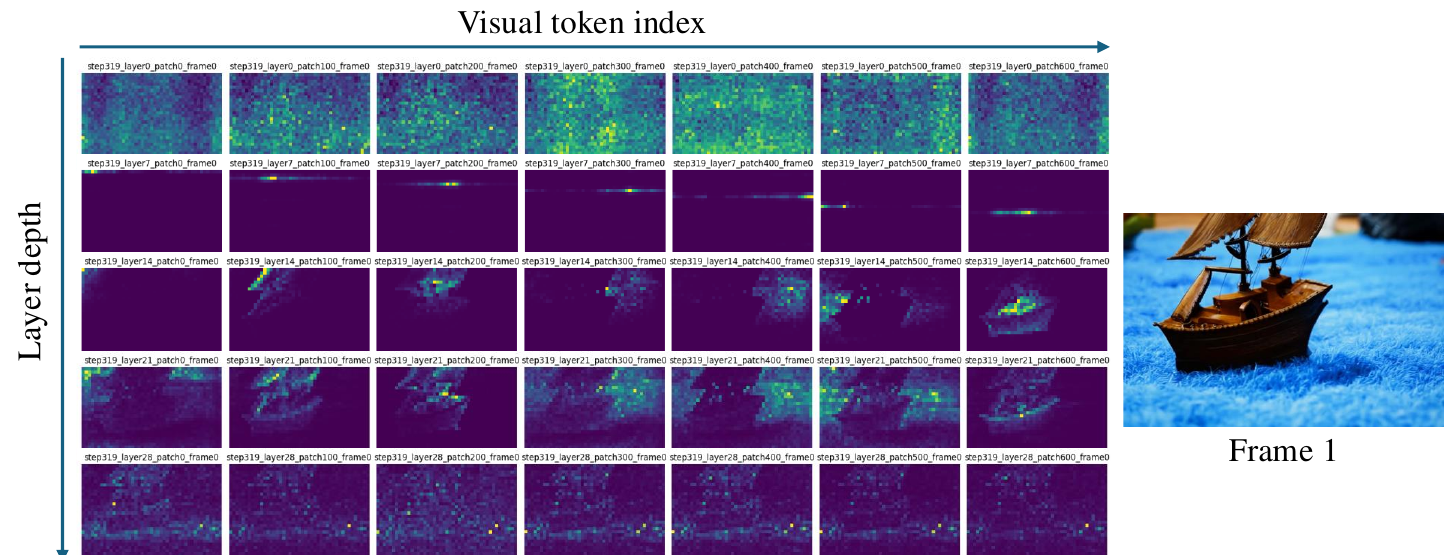}
    \caption{Attention maps between \textit{a visual token} and other visual tokens of the first frame. Some of the maps show similar spatial structure as Frame 1 in the pixel space. The visualization proves that full 3D attention still preserves the spatial structure as in UNet-based diffusion models.}
  \label{fig:appendix_selfattn_map}
\end{figure*}

\begin{figure*}[t]
  \centering
    \includegraphics[width=0.9\linewidth]{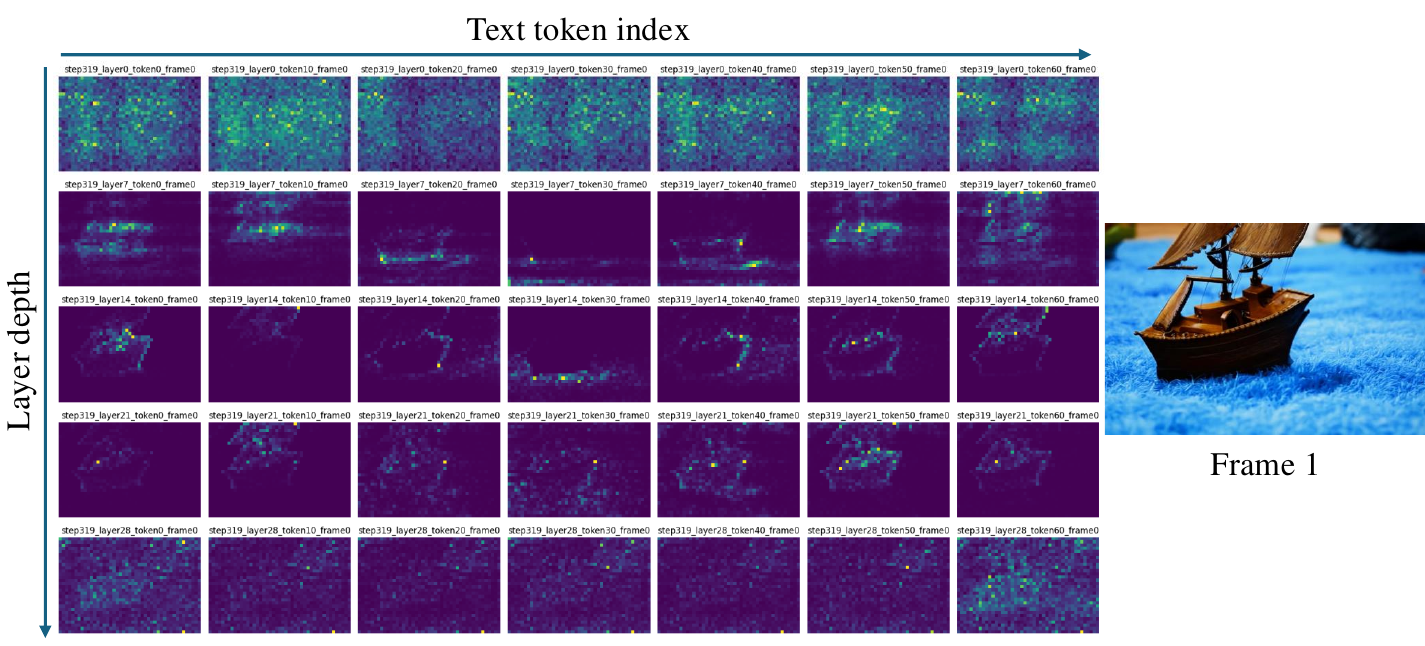}
    \caption{Attention maps between \textit{a text token} and other visual tokens of the first frame. The highlighted regions represent the regions where the text token is highly correlated with. The visualization proves that full 3D attention still preserves the spatial structure as in UNet-based diffusion models.}
  \label{fig:appendix_crossattn_map}
\end{figure*}

\paragraph{DiT attention maps. } While the attention maps from UNet-based image/video diffusion models are shown to reflect the spatial structure of the pixel-space outputs~\cite{hertz2022prompt, ling2024motionclone}, such property in DiT-based video diffusion model~\cite{yang2024cogvideox} with full 3D attention has never been proved, to the best of our knowledge. Here we show that such property still exists in full 3D attentions, which justifies our choice to add masked spatial cross-attention for per-frame context injection. 

In Fig.~\ref{fig:appendix_selfattn_map}, we show the attention maps between \textit{a visual token} and the visual tokens of Frame 1. In Fig.~\ref{fig:appendix_crossattn_map}, we show the attention maps between \textit{a text token} and the visual tokens of Frame 1. Some of the highlighted regions look highly similar as the pixel space structure, proving that even in full 3D attention, the flattened visual tokens still preserves spatial structure. The phenomenon is similar as those observed in UNet-based diffusion models where the spatial and temporal attentions are separated. Please refer to CogVideoX~\cite{yang2024cogvideox} for details of the input and output of the full 3D attentions.

\section{Implementation details}\label{appendix:experiment}
\paragraph{Training data. } For open-domain video generation, our training dataset is obtained by annotating 160K OpenVid~\cite{nan2024openvid} videos, 460K VIDGEN~\cite{tan2024vidgen} videos and 320K videos from HDVILA~\cite{xue2022advancing}. We try to maintain a good balance between human video and non-human videos where the latter outweighs the former as human figures are more challenging to synthesize. While all 940K videos are used for VideoCrafter2-based training, only half of the videos ($\sim$500K) satisfy the length requirement of CogVideoX. Therefore, the training dataset size for \methodname{} based on CogVideoX is effectively $\sim$500K videos. 

For the multi-view scene experiment, we use the ScanNet++ dataset, consisting of 1130 training video clips where each clip has 128 frames. As VideoCrafter2 generates videos of 16 frames, we sample 16 consecutive frames from the 128 frames with a stride of 8. Therefore, we obtain 15 sub-clips with overlaps from each 128-frame video. We prepare 517 16-frame clips for validation purpose and 1392 16-frame clips as the testing set for final evaluation. 

\paragraph{Model Architecture and Training. } We fine-tuned both VideoCrafter2~(VC2)~\cite{chen2024videocrafter2} (U-Net) and CogVideoX-5B~\cite{yang2024cogvideox} (DiT) with our annotated datasets. For VC2 fine-tuning, we add one masked spatial cross-attention in every spatial transformer block and one masked 3D attention after temporal transformer blocks where the latent feature has a spatial resolution less than or equal to 32$\times$32. For CogVideoX, we add one masked spatial cross-attention after every three DiT blocks and one masked 3D self-attention after every six DiT blocks. For open-domain and indoor videos, we fine-tuned VC2 with a learning rate 1e-4 for 20k steps with batch size 256 and 1000 warmup steps. We fine-tuned CogVideoX on open-domain videos with a learning rate 5e-5 for 6k steps. All training processes are done on 64 or 128 NVIDIA A100 GPUs.

\paragraph{Evaluation metrics. } For \textbf{layout-to-video generation evaluation}, we apply the following metrics: FVD, mean Intersection-over-Union~(mIOU), rCLIP$_t$, rCLIP$_i$ and cCFC. To compute mIOU, we first apply Grounding DINO~\cite{liu2023grounding}+SAM2~\cite{ravi2024sam} using the ground truth object labels to obtain object bounding boxes (bboxes) per frame. Then we compute the IOU between the detected bbox in a frame with the ground truth bbox in that frame for the same object. mIOU is the average IOU value over all objects in all involved frames of all videos. If the number of objects from detection and tracking does not match with the number of objects in the ground truth annotation, we keep the most confident detection results up to the number of ground truth bboxes. Then we match each detection bbox to a unique ground truth bbox that produces the highest possible IOU value. For rCLIP$_t$, we crop out regions using the ground truth bboxes. If the region is paired with a blob description, we use CLIP to compute the cosine similarity between the visual region and the blob description. The average similarity score over all videos, all involved frames and all objects give out the rCLIP$_t$ value. rCLIP$_i$ is computed in a similar way but using the bbox region from the ground truth video frame instead of the blob descriptions. It usually has a higher value because the compared features lie in the same output space of CLIP image encoder. As for rCFC, we utilize the detection+tracking results from mIOU and crop out the bbox regions of each object in every frame. Then we compute the cosine similarity between two regions of the same object from two consecutive frames. For one object in a generated video with $T$ frames, this ends up with $T-1$ rCFC values. The reported rCFC is the average value over all detected objects and all videos. 

Note that different methods condition on layouts in different number of frames. Therefore, for a fair comparison, we compute mIOU, rCLIP$_t$, and rCLIP$_i$ only on the frames with the layout condition. For TrackDiffusion~\cite{li2023trackdiffusion}, all 16 generated frames are involved as all frames are grounded on input layouts. For LVD~\cite{lianllm}, Frame 1,  4,  7, 10, 13, 16 of all 16 frames are involved. For VideoTetris~\cite{tian2024videotetris}, Frame 9, 17, 25 of all 32 frames are involved. For \methodname{} based on VC2, we compute the metrics on all 16 frames. For \methodname{} based on CogVideoX, we evaluate on Frame $4k+1$ where $k=0,1,...,12$ out of 49 frames, because there are 13 latent frames due to the 4$\times$ temporal expansion rate from the VAE decoder. 

For \textbf{text-to-video generation evaluation} on T2V-CompBench~\cite{sun2024t2v} and TC-Bench~\cite{feng2024tc}, we adopt the official evaluation metrics. In summary, T2V-CompBench applies different computation methods for different dimension. Consistent Attribute Binding (Consist.-Attr.) and Dynamic Attribute Binding (Dynamic-Attr.) applies LLaVA-v1.6-34B~\cite{liu2024llavanext} to evaluate the attribute correctness. Spatial and Numeracy accuracy are computed using GroundingSAM~\cite{liu2023grounding} to locate and count the objects. Motion Binding is computed using GroundingSAM and Dense Optical Tracking~\cite{lemoing2024dense}. TC-Bench adopts GPT-4 Turbo to answer a list of assertion questions related to compositions of the video. TCR is the percentage (\%) of videos with all assertions passed and TC-Score is the ratio of assertions passed. For details of these metrics, we refer our readers to the original papers~\cite{sun2024t2v, feng2024tc}.

For \textbf{multi-view image generation} in indoor scenes, we compute FID, IS, and CLIP Similarity for image-based metrics, FVD, PSNR, CFC, and rCFC for video-based metrics. CLIP Similarity refer to the average CLIP cosine similarity between each frame and the global caption of the scene. For PSNR, we warp the last frame to an image under current camera view. Then we compute the PSNR between the warped frame and the generated current frame for the regions with content, which reflects a global consistency between two frames. CFC is the average CLIP cosine similarity between any two consecutive frames in the generated videos. rCFC adopts the ground truth annotation and computes the CLIP cosine similarity between two regions of the same object from two consecutive frames. CFC and rCFC reflects video consistency in different granularity levels. 

\paragraph{In-context learning examples. } We show the full prompt and our in-context exemplars in Table~\ref{tab:appendix_icl}. The layouts follow a JSON format which allows GPT-4o to produce outputs that can be robustly parsed by a JSON parser. We use two fixed exemplars for all prompts in our text-to-video generation experiments. While this simplest in-context design can lead to many flaws in the generated layouts, our pipeline of GPT-4o+\methodname{} still demonstrates strong performances in many compositional aspects, suggesting great potential in further improving the performance by more sophisticated layout generation approaches.

\section{Additional Results}\label{appendix:qualitative}

\paragraph{Ablation study. } In Table~\ref{tab:appendix_ablation}, we show the ablation study on context interpolation methods. We emphasize the importance of context interpolation by comparing row 1 with other rows. For ``None'' interpolation method, we simply use empty strings for frames without blob descriptions and obtain context features from CLIP text encoder. Note that this has led to apparent performance drop in all metrics compared to using simple linear interpolation in row 2. Therefore, the existence of interpolation for context features is essential to generate consistent videos and enhance prompt-video alignment. 

\begin{table}
  \centering
  \resizebox{\linewidth}{!}{%
  \begin{tabular}{l ccc| ccccc}
    \toprule
    & \shortstack{Mask \\3D Attn} & \shortstack{Context\\Interp.} & \shortstack{Training\\Data} & FVD $\downarrow$ & mIOU $\uparrow$ & rCLIP$_t$ $\uparrow$ & rCLIP$_i$ $\uparrow$ & rCFC $\uparrow$ \\
    \midrule
    \rowcolor[gray]{0.8}
    \texttt{1} & $\checkmark$ & None & 400K & 379 & 0.5614 & 0.2767 & 0.8161 & 0.9466 \\
    \texttt{2} & $\checkmark$ & Linear & 400K & \textbf{346} & 0.5771 & 0.2794 & 0.8200 & \textbf{0.9480} \\
    \texttt{3} & $\checkmark$ & Slerp & 400K  & 378 & 0.5702 & 0.2763 & 0.8142 & 0.9438 \\
    \texttt{4} & $\checkmark$ & Perceiver & 400K & 352 & \textbf{0.5926} & \textbf{0.2806} & \textbf{0.8204} & 0.9459 \\
    \bottomrule
  \end{tabular}
  }
  \caption{Ablation study on model architecture, context interpolation method and training data size on YTVIS-700. In the highlighted row, no interpolation method is applied. For frames without blob descriptions, we input empty strings to CLIP text encoder to get context features. We can see a consistent performance drop without the context interpolation, as indicated by all five metrics.
  }
  \label{tab:appendix_ablation}
\end{table}

\subsection{Additional qualitative results}
\begin{figure*}[t]
  \centering
    \includegraphics[width=\textwidth]{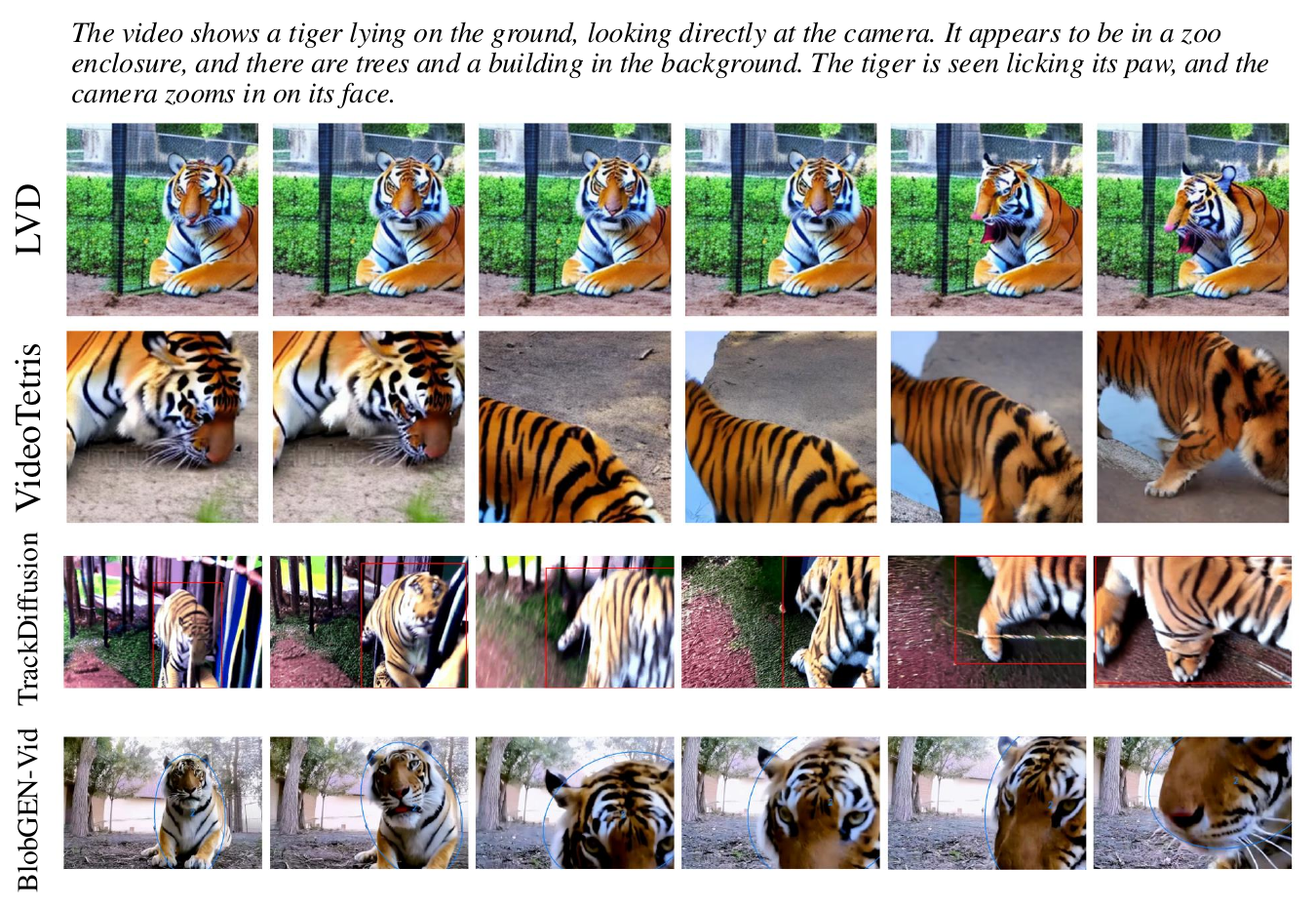}
    \caption{Qualitative examples from YoutubeVIS-700}
  \label{fig:appendix_qualitative1}
\end{figure*}

\begin{figure*}[t]
  \centering
    \includegraphics[width=\textwidth]{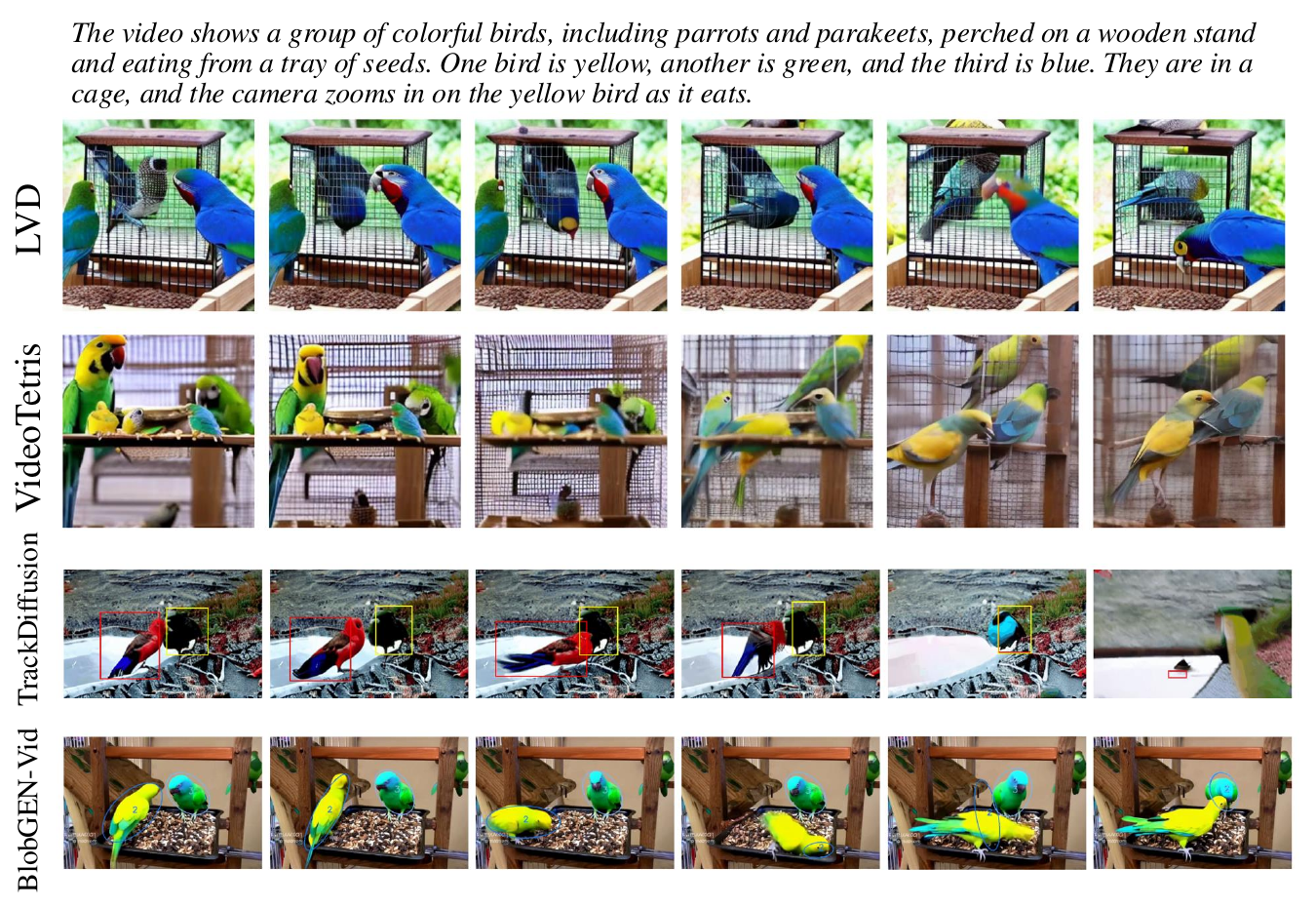}
    \caption{Qualitative examples from YoutubeVIS-700}
  \label{fig:appendix_qualitative2}
\end{figure*}

\begin{figure*}[t]
  \centering
    \includegraphics[width=\textwidth]{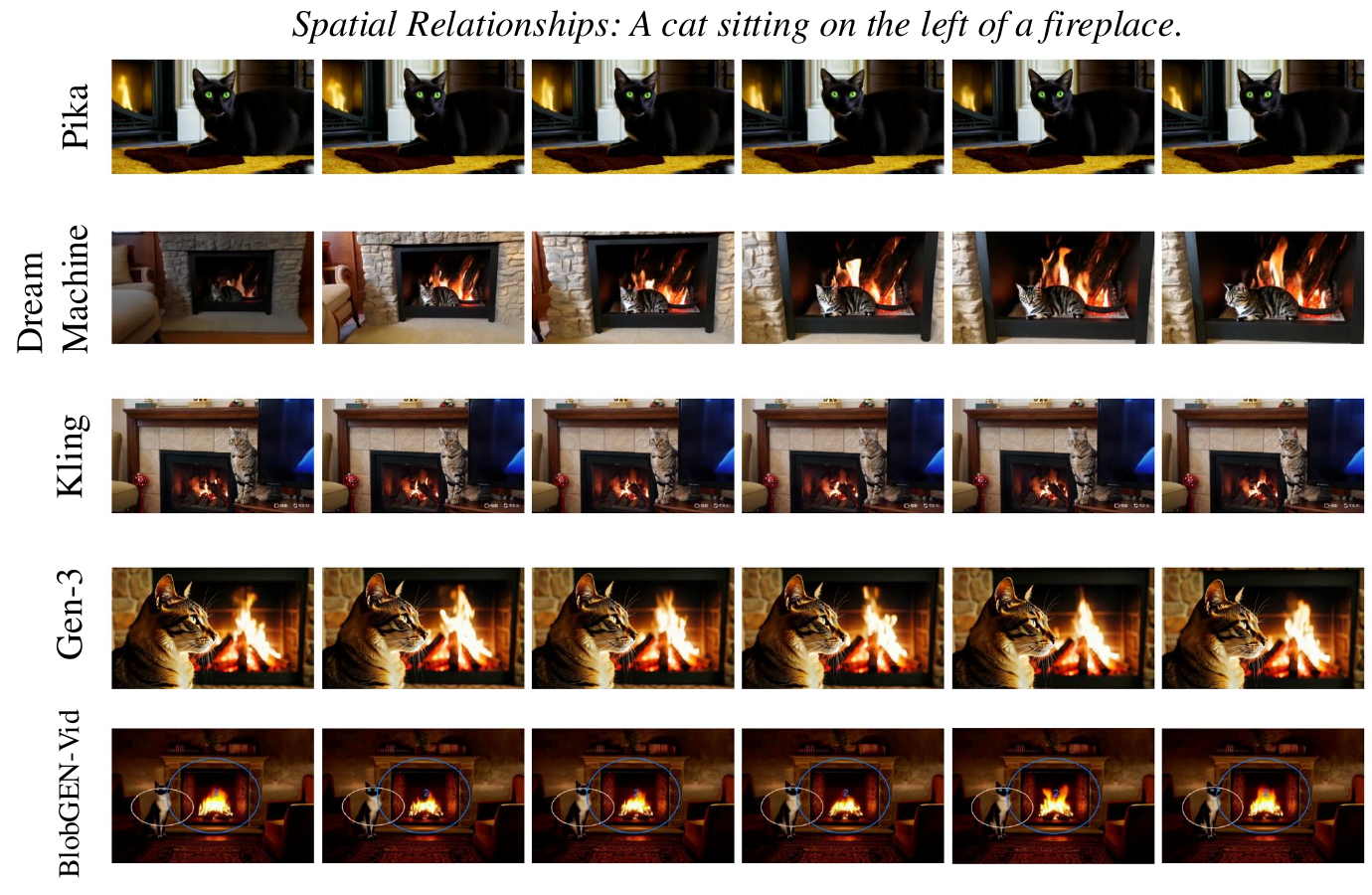}
    \caption{Qualitative examples from T2V-CompBench}
  \label{fig:appendix_qualitative3}
\end{figure*}

\begin{figure*}[t]
  \centering
    \includegraphics[width=\textwidth]{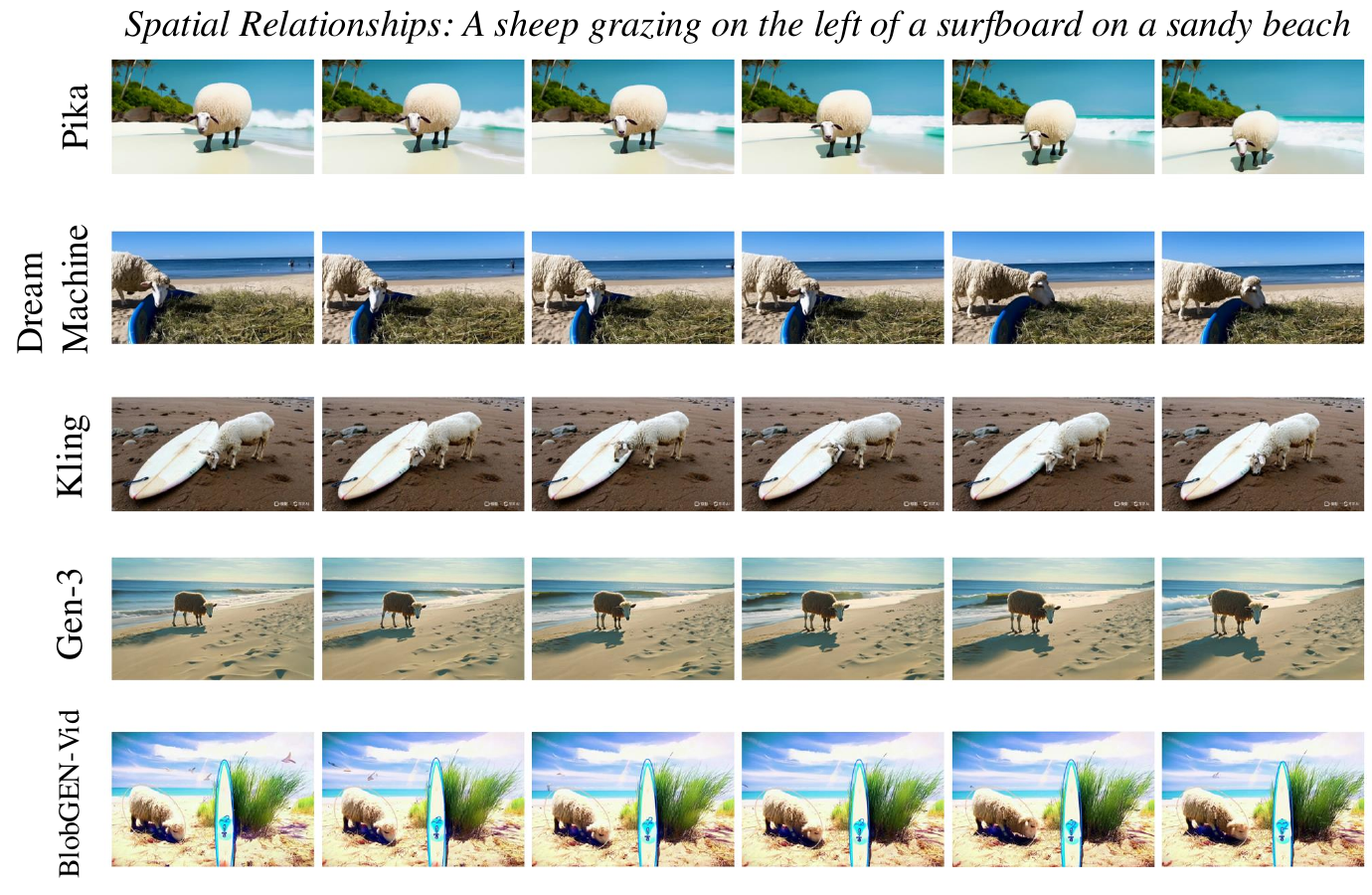}
    \caption{Qualitative examples from T2V-CompBench}
  \label{fig:appendix_qualitative4}
\end{figure*}

\begin{figure*}[t]
  \centering
    \includegraphics[width=\textwidth]{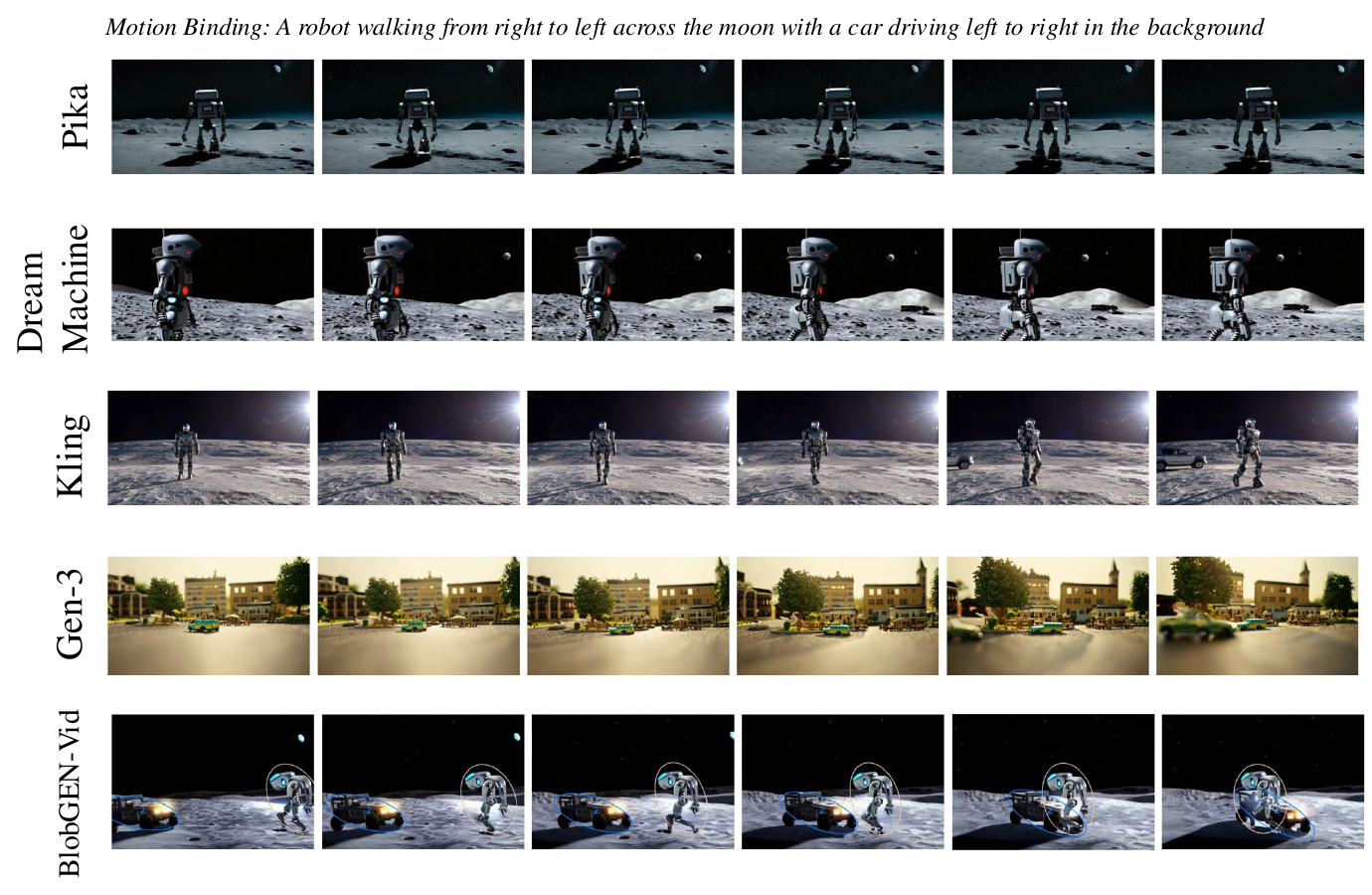}
    \caption{Qualitative examples from T2V-CompBench}
  \label{fig:appendix_qualitative5}
\end{figure*}

\begin{figure*}[t]
  \centering
    \includegraphics[width=\textwidth]{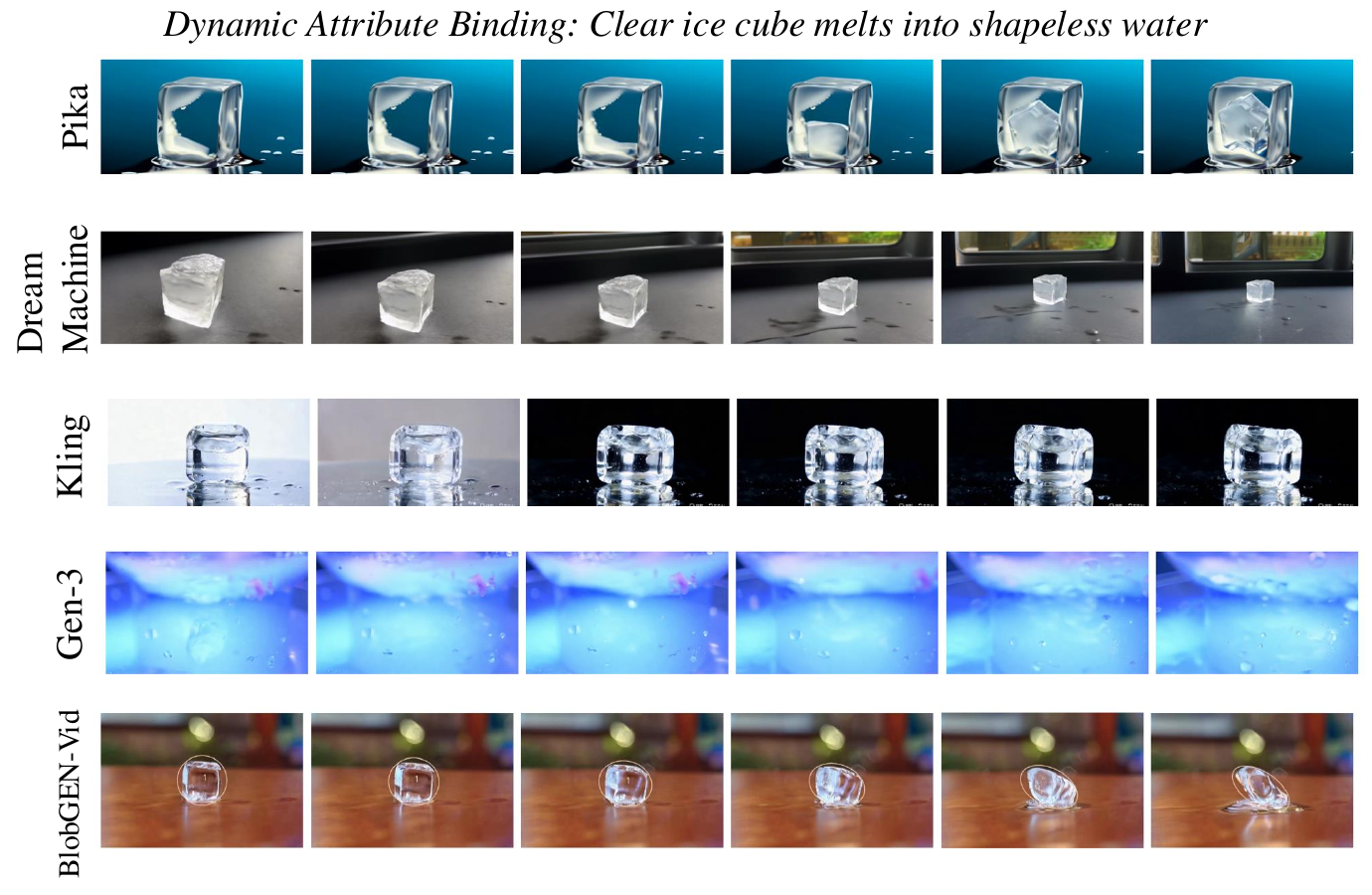}
    \caption{Qualitative examples from T2V-CompBench}
  \label{fig:appendix_qualitative6}
\end{figure*}

\begin{figure*}[t]
  \centering
    \includegraphics[width=\textwidth]{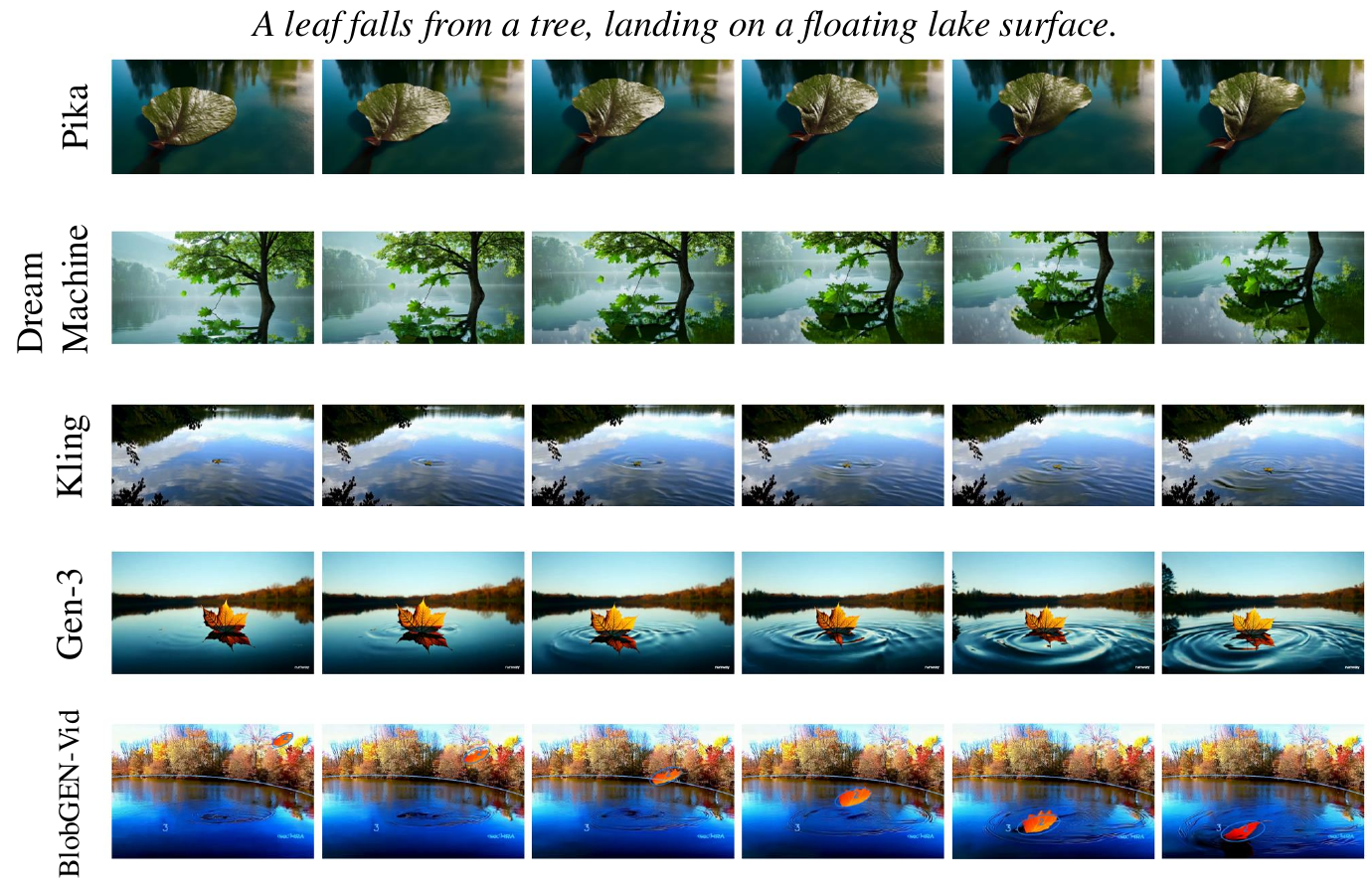}
    \caption{Qualitative examples from TC-Bench}
  \label{fig:appendix_qualitative7}
\end{figure*}

\begin{figure*}[t]
  \centering
    \includegraphics[width=\textwidth]{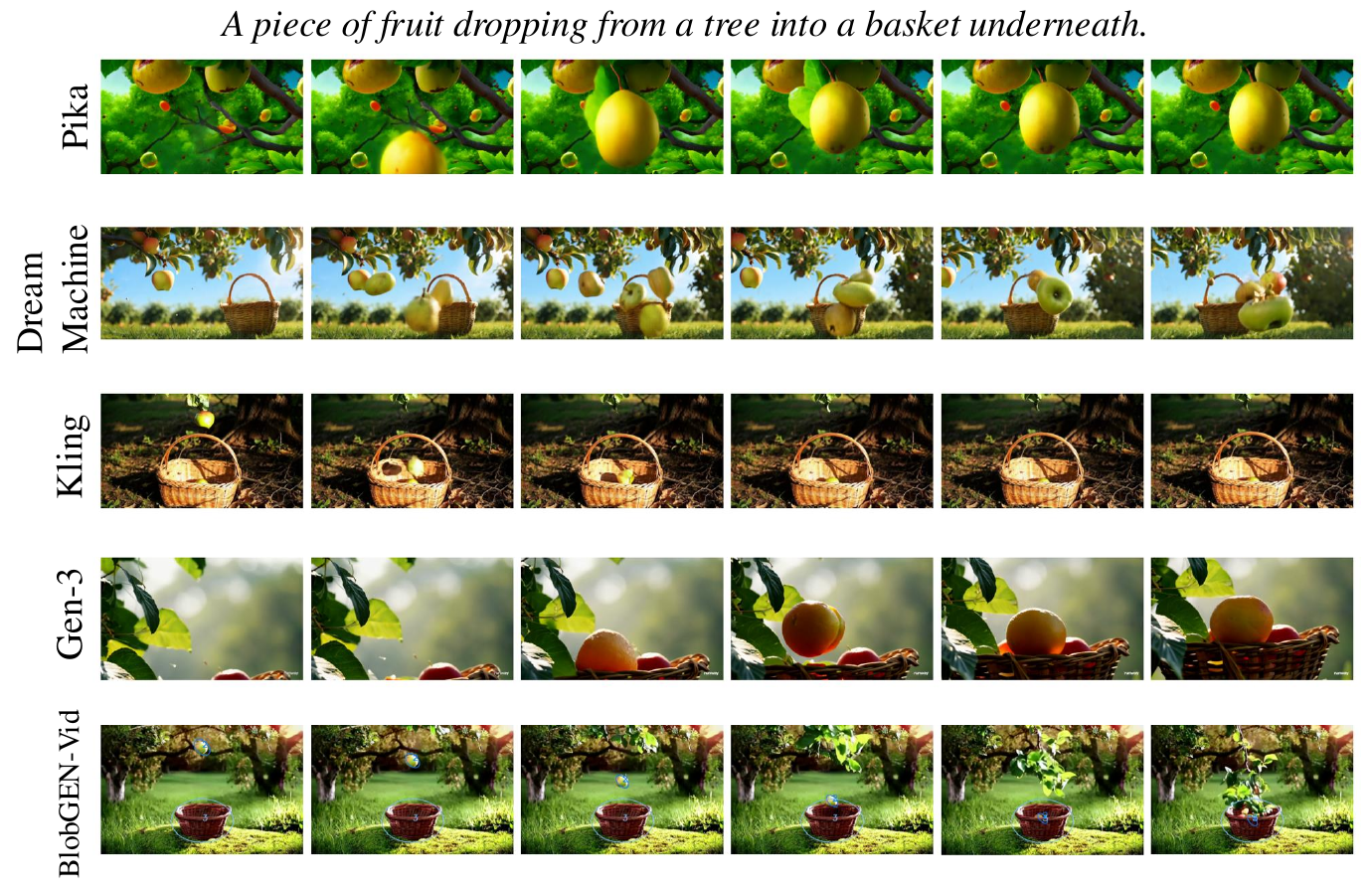}
    \caption{Qualitative examples from TC-Bench}
  \label{fig:appendix_qualitative8}
\end{figure*}

\begin{figure*}[t]
  \centering
    \includegraphics[width=\textwidth]{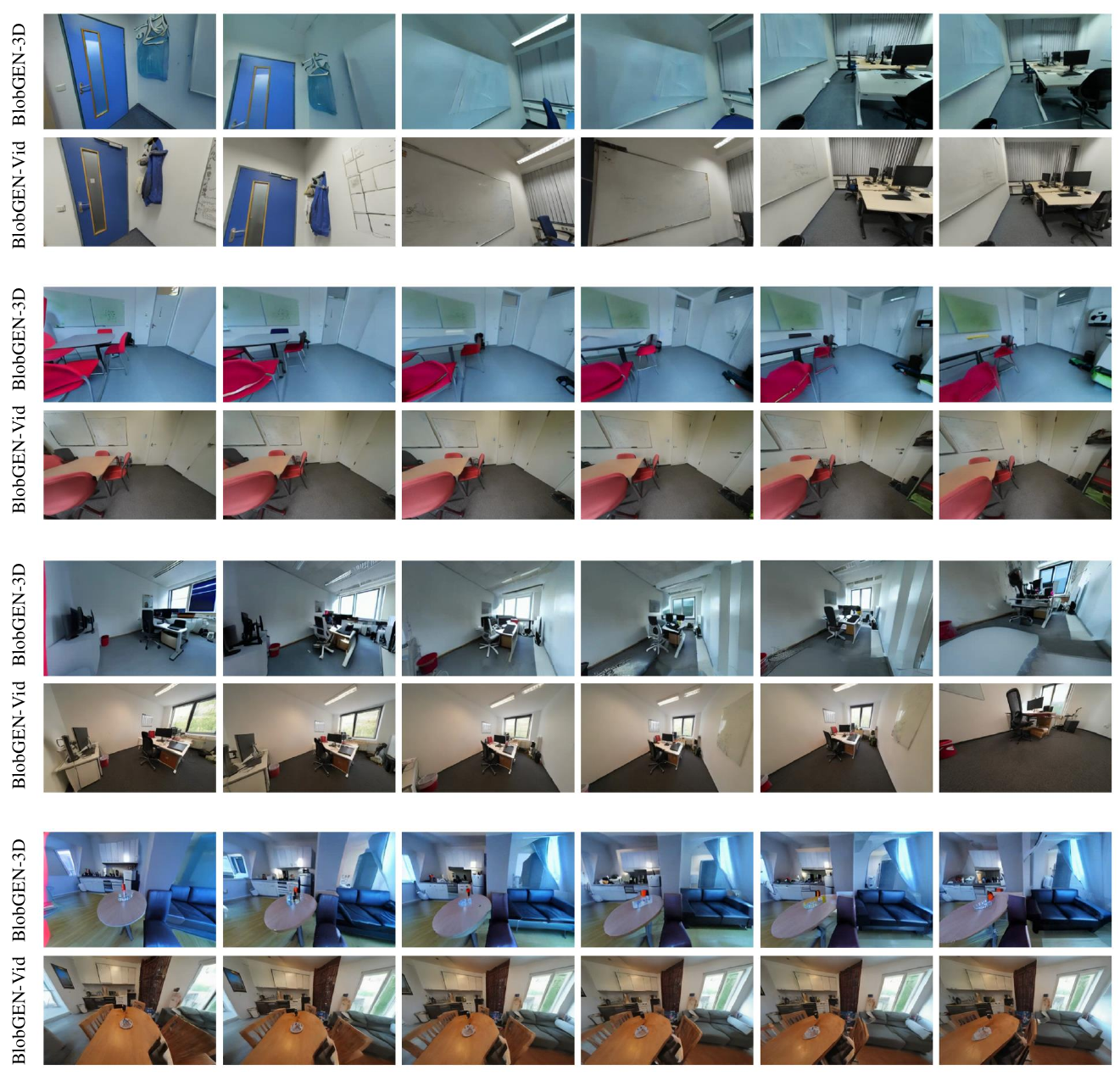}
    \caption{Qualitative examples from ScanNet++}
  \label{fig:appendix_qualitative9}
\end{figure*}

We show additional qualitative results from various settings and benchmarks in Fig.~\ref{fig:appendix_qualitative1}-\ref{fig:appendix_qualitative9}.

\begin{table*}[h]
    \centering
    \begin{adjustbox}{width=.8\textwidth, center}
    \begin{tabular}{@{}p{16cm}@{}}
    \toprule
    Generate a video layout using ellipses for the given user prompt. Each ellipse should be represented with five parameters and a paired object caption. The parameters are [cx, cy, a, b, theta] where cx and cy are the center coordinates, a and b are the major and minor axes length, and theta is the rotation angle. Assume there are 13 frames in the video, and you should generate layouts for Frame0,2,4,...,12. The video resolution is 720 width and 480 height. Try to cover all objects mentioned in the prompt. You should follow the format of the following examples:
    \newline\newline
    Example 1:\newline
    Prompt: The video shows a small owl perched on a branch, looking around. It appears to be in a natural habitat, surrounded by greenery. The owl is alert and focused, possibly observing its surroundings or looking for prey. The camera angle is from below, giving a clear view of the owl's feathers and features.
    \newline
    \texttt{```}json
    \newline
    {``Frame0'': {``Object2'': {``blob'': [443, 252, 102, 72, -2.353], \newline
       ``caption'': ``The bird in the close-up image is a small, brown creature with a white belly. It appears to be in mid-flight, with its wings spread wide and its tail fanned out. The bird is perched on a tree branch, which is covered in green leaves. The bird``s eyes are open, and it seems to be looking directly at the camera. ''}}, \newline
     ``Frame2'': {``Object2'': {``blob'': [438, 253, 106, 68, -2.357],\newline
       ``caption'': `` The bird in the close-up image is a small, brown and white bird with a prominent beak, perched on a tree branch. The bird turns its head to the side.''}}, \newline
        ... \newline
     ``Frame12'': {``Object2'': {``blob'': [445, 249, 119, 57, -2.023],\newline
       ``caption'': `` The bird in the close-up image is a small owl perched on a tree branch. The bird is looking upwards, turning its face away from the camera. ''}}} \newline
    \texttt{```}
    \newline
    \newline
    Example 2:\newline
    Prompt: The video shows a woman leading a horse while a young girl rides on its back. The girl is wearing a helmet and a riding jacket, and the woman is holding the reins. They are in a stable or a similar outdoor area with several parked cars in the background.\newline
    \texttt{```}json\newline
    {``Frame0'': {``Object2'': {``blob'': [365, 277, 93, 64, 1.749], \newline
       ``caption'': ``The horse in the close-up image is a small, brown pony. It is wearing a saddle and a bridle, indicating it is prepared for riding. The pony appears to be walking on a street, with a red car visible in the background. ''},\newline
      ``Object3'': {``blob'': [165, 247, 102, 75, -3.095], \newline
       ``caption'': ``The car in the close-up image is a black Volkswagen Beetle. It has a distinctive rounded shape and a yellow license plate. The car appears to be in motion on a road. ''},\newline
      ``Object4'': {``blob'': [563, 276, 132, 44, 1.599], \newline
       ``caption'': ``The image is blurry, making it difficult to discern specific details about the person. The person appears to be walking, possibly in a parking lot or similar outdoor setting. The individual is holding onto a leash, suggesting they might be walking a dog. ''}},\newline
     ...\newline
     ``Frame12'': {``Object2'': {``blob'': [387, 229, 136, 95, 2.685], \newline
       ``caption'': `` The horse in the close-up image is a large, brown horse with a white blaze on its face. It appears to be a healthy and well-groomed animal. ''},\newline
      ``Object3'': {``blob'': [30, 188, 87, 70, -2.388], \newline
       ``caption'': `` The car in the close-up image is a black sedan with a yellow license plate. The vehicle appears to be parked or stationary, as indicated by the lack of motion blur. ''},\newline
      ``Object4'': {``blob'': [670, 242, 151, 64, 1.598], \newline
       ``caption'': `` The image is a close-up of a person who appears to be a woman. She is holding a leash, which suggests she might be with a pet. The woman is wearing a white top and blue jeans. ''}}}\newline
    \texttt{```}
    \newline\newline
    Prompt: \{inference prompt\} 
    \\\bottomrule
    \end{tabular}
    \end{adjustbox}
    \caption{Our prompt for GPT-4o to generate blob layouts in text-to-video generation. We use two fixed exemplars for all prompts as shown in this table. The ``\{inference prompt\}'' represents the actual text prompt that users use to generate a video.}
    \label{tab:appendix_icl}
\end{table*}

\end{document}